%% file: draft.tex
\documentclass[final,1p,times,authoryear]{elsarticle}

\usepackage{graphicx,float,psfrag,amsmath,amsthm,amssymb,nicefrac}
\usepackage{algorithm,algpseudocode}
\usepackage{enumitem}
\usepackage{stfloats}

\journal{Cognitive Systems Research}

\begin{document}

\begin{frontmatter}



\title{Learning sparse representations in reinforcement learning}

\author{Jacob Rafati}
\ead{jrafatiheravi@ucmerced.edu}
\ead[url]{http://rafati.net/}
\author{David C. Noelle}
\ead{dnoelle@ucmerced.edu}

\address{Electrical Engineering and Computer Scinence\\
	Computational Cognitive Neuroscience Laboratory\\
	University of California, Merced\\
	5200 North Lake Road, Merced, CA 95343 USA.}

\begin{abstract}
Reinforcement learning (RL) algorithms allow artificial agents to improve their selection of actions to increase rewarding experiences in their environments. Temporal Difference (TD) Learning -- a model-free RL method -- is a leading account of the midbrain dopamine system and the basal ganglia in reinforcement learning. These algorithms typically learn a mapping from the agent's current sensed state to a selected action (known as a \emph{policy} function) via learning a value function (expected future rewards). TD Learning methods have been very successful on a broad range of control tasks, but learning can become intractably slow as the state space of the environment grows. This has motivated methods that learn internal representations of the agent's state, effectively reducing the size of the state space and restructuring state representations in order to support generalization. However, TD Learning coupled with an artificial neural network, as a function approximator, has been shown to fail to learn some fairly simple control tasks, challenging this explanation of reward-based learning. We hypothesize that such failures do not arise in the brain because of the ubiquitous presence of lateral inhibition in the cortex, producing sparse distributed internal representations that support the learning of expected future reward. The sparse conjunctive representations can avoid catastrophic interference while still supporting generalization. We provide support for this conjecture through computational simulations, demonstrating the benefits of learned sparse representations for three problematic classic control tasks: Puddle-world, Mountain-car, and Acrobot. 
\end{abstract}


\end{frontmatter}


\section{Introduction}
\label{sec:intro}
\input{sections/introduction.tex}

\section{Reinforcement learning}
\label{sec:rl}
\input{sections/background.tex}

\section{Methods for learning sparse representations}
\label{sec:kwta}
\input{sections/sparse.tex}

\section{Experiments and Simulation Tasks}
\label{sec:simulations}
\input{sections/simulation-tasks.tex}

\section{Results and discussions}
\label{sec:results}
\input{sections/results.tex}

\section{Conclusions and Future Work}
\label{sec:conclusions}
\input{sections/conclusions.tex}



\bibliographystyle{elsarticle-harv} 
\bibliography{refs.bib}




\end{document}

%% file: sections/introduction.tex
Reinforcement learning (RL) -- a class of machine learning problems -- is learning how to map situations to actions so as to maximize numerical reward signals received during the experiences that an artificial agent has as it interacts with its environment \citep{Sutton:Barto:1998:RL-BOOK-1}. An RL agent must be able to sense the state of its environment and must be able to take actions that affect the state. The agent may also be seen as having a goal (or goals) related to the state of the environment.

Humans and non-human animals' capability of learning highly complex
skills by reinforcing appropriate behaviors with reward and the role of midbrain dopamine system in reward-based
learning has been well described by a class of a model-free RL, called \emph{Temporal Difference (TD) 	Learning}~\citep{MontaguePR:1996:Dopamine,SchultzW:1997:Science}. While
TD Learning, by itself, certainly does not explain all observed
RL phenomena, increasing evidence suggests that it
is key to the brain's adaptive nature~\citep{DayanP:2008:Ugly}.

One of the challenges that arise in RL in real-world problems is that the state space can be very large. This is a version of what has classically been called the \emph{curse of dimensionality}. Non-linear function approximators coupled with reinforcement learning have made it possible to learn abstractions over high dimensional state spaces. Formally, this function approximator is a parameterized equation that maps from state to value, where the parameters can be constructively optimized based on the experiences of the agent. One common function approximator is an artificial neural network, with the parameters being the connection weights in the network. Choosing a right structure for the value function approximator, as well as a proper method for learning representations are crucial for a robust and successful learning in TD    \citep{Rafati2019phd,Rafati-Marcia:2019:RLG-arXiv,Rafati-Noelle:2019:HRL-arXiv}.

Successful examples of using neural networks for RL include learning how to play the game of Backgammon at the Grand Master level~\citep{TesauroG:1995:TDGammon}. Also, recently, researchers at DeepMind Technologies used deep convolutional neural networks (CNNs) to learn how to play some ATARI games from raw video data \citep{DeepMind:Nature:2015}. The resulting performance on the games was frequently at or better than the human expert level. In another effort, DeepMind used deep CNNs and a Monte Carlo Tree Search algorithm that combines supervised learning and reinforcement learning to learn how to play the game of Go at a super-human level \citep{DeepMind-AlphaGo}. 

\subsection{Motivation for the research}
Despite these successful examples, surprisingly, some relatively
simple problems for which TD coupled with a neural network function approximator has been shown to fail. For example, learning to navigate to a goal location in a simple two-dimensional space (see Figure \ref{fig:puddle-world-env}) in which there are obstacles has been shown to pose a substantial challenge to TD Learning using a backpropagation neural network \citep{Boyan:1995:Approximating}. Note that the proofs of convergence to optimal performance depend on the agent maintaining a potentially highly discontinuous value function in the form of a large
look-up table, so the use of a function approximator for the value
function violates the assumptions of those formal analyses. Still, it
seems unusual that this approach to learning can succeed at some
difficult tasks but fail at some fairly easy tasks.

The power of TD Learning to explain biological RL
is greatly reduced by this observation. If TD Learning fails at simple
tasks that are well within the reach of humans and non-human animals,
then it cannot be used to explain how the dopamine system supports
such learning.

In response to \cite{Boyan:1995:Approximating}, \cite{SuttonRS:1996:Coarse} showed that a TD Learning agent can learn this task by hard-wiring the hidden
layer units of the backpropagation network (used to learn the value
function) to implement a fixed sparse conjunctive (coarse) code of the
agent's location. The specific encoding used was one that had been
previously proposed in the CMAC model of the
cerebellum~\citep{AlbusJS:1975:CMAC}. Each hidden unit would become
active only when the agent was in a location within a small region. For any given location, only a small fraction of the hidden units displayed non-zero activity. This is what it means for the hidden representation to be a ``sparse'' code. Locations that were close to each other in the environment produced more overlap in the hidden units that were active than
locations that were separated by a large distance. By ensuring that
most hidden units had zero activity when connection weights were
changed, this approach kept changes to the value function in one
location from having a broad impact on the expected future reward at
distant locations. By engineering the hidden layer
representation, this RL problem was solved.

This is not a general solution, however. If the same approach was
taken for another RL problem, it is quite possible
that the CMAC representation would not be appropriate. Thus, the
method proposed by \cite{SuttonRS:1996:Coarse} does not help us
understand how TD Learning might flexibly learn a variety of
RL tasks. This approach requires prior knowledge
of the kinds of internal representations of sensory state that are
easily associated with expected future reward, and there are simple
learning problems for which such prior knowledge is unavailable.

We hypothesize that the key feature of the
\cite{SuttonRS:1996:Coarse} approach is that it produces a sparse
conjunctive code of the sensory state. Representations of this kind
need not be fixed, however, but might be learned at the hidden layers
of neural networks. 

There is substantial evidence that sparse representations are generated in the cortex by neurons that release the transmitter GABA \citep{OReillyRC:2001:CECN} via lateral inhibition. Biologically inspired models of the brain show that, the sparse representation in the hippocampus can minimize the overlap of representations assigned to different cortical patterns. This leads to \emph{pattern separation}, avoiding the catastrophic interference, but also supports \emph{generalization} by modifying the synaptic connections so that these representations can later participate jointly in \emph{pattern completion} \citep{O'Reilly-and-McClelland:1994:Hippocampus,Noelle-2008:BICA}. 

Computational cognitive neuroscience models have
shown that a combination of feedforward and feedback inhibition
naturally produces sparse conjunctive codes over a collection of
excitatory neurons \citep{OReillyRC:2001:CECN}. Such patterns of
lateral inhibition are ubiquitous in the mammalian
cortex \citep{KandelE:2012:Book}. Importantly, neural networks containing such lateral inhibition can still learn to represent input information in
different ways for different tasks, retaining flexibility while producing the kind of sparse conjunctive codes that may support reinforcement learning. Sparse distributed representation schemes have the useful properties of coarse codes while reducing the likelihood of interference between different representations.

\subsection{Objective of the paper}
In this paper, we demonstrate how incorporating a ubiquitous feature
of biological neural networks into the artificial neural networks used
to approximate the value function can allow TD Learning to succeed at
simple tasks that have previously challenged it. Specifically, we show
that the incorporation of \emph{lateral inhibition}, producing
competition between neurons so as to produce \emph{sparse conjunctive
	representations}, can produce success in learning to approximate the
value function using an artificial neural network, where only failure
had been previously found. Thus, through computational simulation, we
provide preliminary evidence that lateral inhibition may help
compensate for a weakness of TD Learning, improving this machine
learning method and further buttressing the TD Learning account of
dopamine-based reinforcement learning in the brain. This paper extends our previous works, \cite{Rafati-Noelle:2015:CogSci,Rafati-Noelle:2017:CCN}.

\subsection{Outline of the paper}
The organization of this paper is as follows. In Section \ref{sec:rl}, we provide background on the reinforcement learning problem and the temporal difference learning methods. In Section \ref{sec:kwta}, we introduce a method for learning sparse representation in reinforcement learning inspired by the lateral inhibition in the cortex. In Section \ref{sec:simulations}, we provide details concerning our computational simulations of TD Learning with lateral inhibition to solve some simple tasks that TD methods were reported to fail to learn in the literature. In Section \ref{sec:results}, we present the results of these simulations and compare the performance of our approach to previously examined methods. The concluding remarks and future research plan can be found in Section \ref{sec:conclusions}.

%% file: sections/background.tex
\subsection{Reinforcement learning problem}
\begin{figure}
	\centering
	\includegraphics[width=0.7\textwidth]{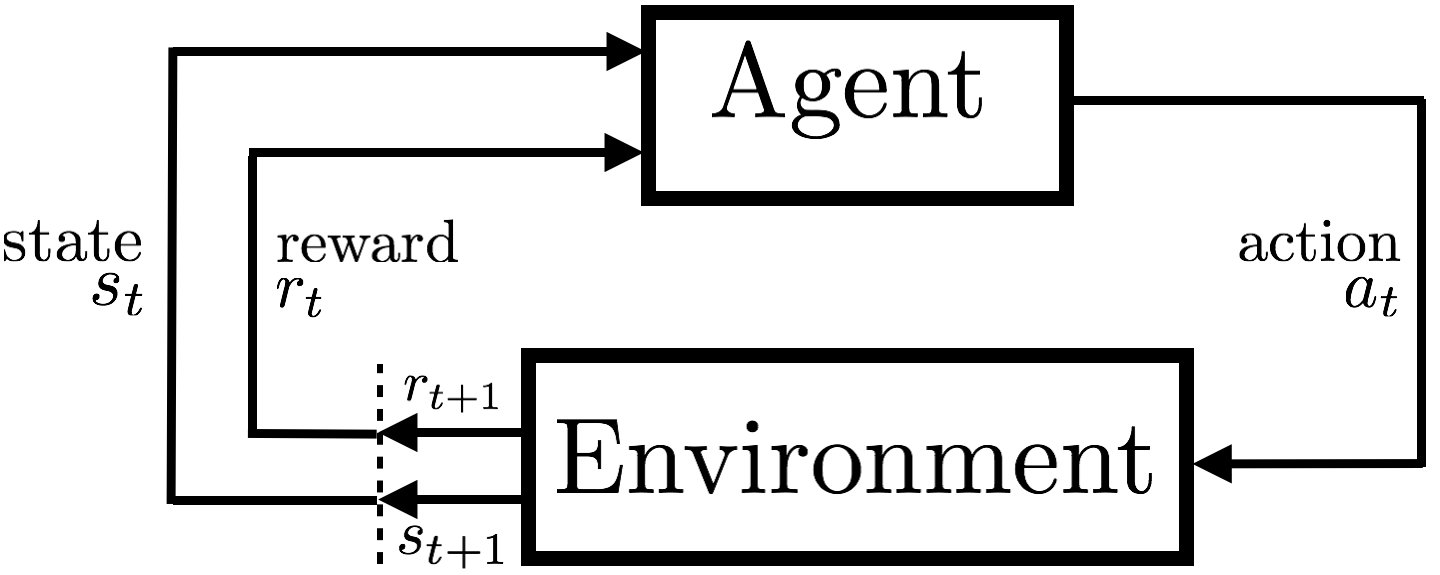}
	\caption{The agent/environment interaction in reinforcement learning \citep{Sutton:Barto:2017:RL-BOOK-2}.}
	\label{fig:agent-env}
\end{figure}

The Reinforcement Learning (RL) problem is learning through interaction with an \emph{environment} to achieve a \emph{goal}. The learner and decision maker is called the \emph{agent}, and everything outside of the agent is called the \emph{environment}. The agent and the environment interact over a sequence of discrete time steps, $t = 0,1,2,\dots$. At each time step, $t$, the agent receives a representation of the environment's \emph{state}, $S_t \in \mathcal{S}$, where $\mathcal{S}$ is the set of all possible states, and on that basis the agent selects an \emph{action}, $A_t \in \mathcal{A}$, where $\mathcal{A}$ is the set of all possible actions for the agent. One time step later, at $t+1$, as a consequence of the agent's action, the agent receives a \emph{reward} $R_{t+1} \in \mathbb{R}$ and also an update on the agent's new state, $S_{t+1}$, from the environment. Each cycle of interaction is called an \emph{experience}. Figure~\ref{fig:agent-env} summarizes the agent/environment interaction (see \citep{Sutton:Barto:2017:RL-BOOK-2} for more details).  

At each time step $t$, the agent implements a mapping from states to possible actions, $\pi_t : \mathcal{S} \rightarrow \mathcal{A}$. This mapping is called the agent's \emph{policy} function. 

The objective of the RL problem is to maximize the expected value of \emph{return}, i.e. the cumulative sum of the received reward, defined as follows
\begin{align}
G_t \triangleq  \sum_{t' = t + 1}^{T} \gamma^{t' - t - 1} r_{t'}, \quad 0 \leq t < T,
\label{eq:def-return}
\end{align}
where $T$ is a final step (also known as horizon) and $0 \leq \gamma \leq 1$ is a discount factor. As $\gamma$ gets closer to $1$, the objective takes future rewards into account more strongly, and if $\gamma$ is closer to $0$, the agent is only concerned about maximizing the immediate reward. The goal of RL problem is finding an optimal policy, $\pi^*$ that maximizes the expected return for each state $s \in \mathcal{S}$
\begin{align}
\pi^*(s) = \arg\max\limits_{\pi} \mathbb{E}_{\pi} [ G_t~|~S_t = s ],
\label{eq:RL--optimization-probelm}
\end{align}
where $\mathbb{E}_{\pi}[.]$ denotes the expected value given that the agent follows policy $\pi$. The reinforcement learning algorithms often involve the estimation of a \emph{value} function that estimate how good it is to take a given action in a given state. 

The value of taking action $a$ under the policy $\pi$ in state $s$ is defined as the expectation of the return starting from state $s$ and taking action $a$, and then following the policy $\pi$ 
\begin{align}
Q_{\pi}(s,a) \triangleq \mathbb{E}_{\pi} [\, G_t ~|~ S_t = s, A_t = a ].
\end{align}

The agent/environment interaction can be broken into subsequences, which we call \emph{episodes}, such as plays of a game or any sort of repeated interaction. Each episode ends when time is over, i.e., $t = T$ or when the agent reaches an absorbing \emph{terminal} state.

\subsection{Generalization using neural network}
The reinforcement learning algorithms  need to maintain an estimate of the value function $Q_{\pi}(s,a)$, and this function could be stored as a simple look-up table. However, when the state space is large or not all states are observable, storing these estimated values in a table is no longer possible. We can, instead, use a function approximator to represent the mapping to the estimated value. A parameterized functional form trainable parameters $w$ can be used: $q(s,a;w) \approx Q_{\pi}(s,a)$. In this case, RL learning requires a search through the space of parameter values, $w$, for the function approximator. One common way to approximate the value function is to use a nonlinear functional form such as that embodied by an artificial neural network. When a change is made to the weights $w$ based on an experience from a particular state, the changed weights will then affect the value function estimates for similar states, producing a form of generalization. Such generalization makes the learning process potentially much more powerful.

\subsection{Temporal difference learning}
Temporal Difference (TD) learning \citep{SuttonRS:1988:TD} is a model-free reinforcement learning algorithm that attempts to learn a policy without learning a model of the environment. TD is a combination of Monte Carlo random sampling and dynamic programming ideas \citep{Sutton:Barto:1998:RL-BOOK-1}. The goal of the TD approach is to learn to predict the value of a given state based on what happens in the next state by bootstrapping, i.e., updating values based on the learned values, without waiting for a final outcome. 

SARSA (State-Action-Reward-State-Action) is an \emph{on-policy} TD algorithm that learns the action-value function \citep{Sutton:Barto:1998:RL-BOOK-1}. Following the SARSA version of TD Learning (see Algorithm \ref{Algo:SARSA}), the reinforcement learning agent is controlled in the following way. The current state of the
agent, $s$, is provided as input to the neural network, producing
output as state-action values $q(s,a;w) \approx Q_{\pi}(s,a)$. The action $a$ is selected using the exploration policy, $\epsilon$-greedy, where with a small exploration probability, $\epsilon$, these values were ignored, and an action was selected uniformly at random from the possible actions $a_i \in \mathcal{A}$, otherwise, the output unit with the highest activation level determined the action, $a$, to be taken, i.e., 
\begin{align}
a = \begin{cases}
\arg\max\limits_a Q(s,a) &\textrm{with prob. of }(1-\epsilon), \\
\textrm{random action from } \mathcal{A}  &\textrm{with prob. of }\epsilon.
\end{cases}
\label{eq:eps-greedy}
\end{align}

The agent takes action $a$ and then, the environment updates the agent's state to $s'$ and the agent received a reward signal, $r$, based on its current state, $s'$. The action selection process was then repeated at state $s'$, determining a subsequent action, $a'$. Before this action was taken, however, the neural network value function approximator had its
connection weights updated according to the SARSA Temporal Difference
(TD) Error:
\begin{align} \delta \ = \ r \ + \ \gamma \ q(s',a';w) -
q(s,a;w).
\end{align}
The TD Error, $\delta$, was used to construct an error signal for the
backpropagation network implementing the value function. The network
was given the input corresponding to $s$, and activation was
propagated through the network. Each output unit then received an
error value. This error was set to zero for all output units except
for the unit corresponding to the action that was taken, $a$. The 
selected action unit received an error signal equal to the TD error,
$\delta$. This error value was then backpropagated through the
network, using the standard backpropagation of error
algorithm~\citep{RumelhartDE:1986:BP}, and connection weights were
updated. Assuming that the loss function is sum of square error (SSE) of the TD error, i.e. $L(w) \triangleq \delta^2/2$, the weights will be updated based on the gradient decent method as $w \gets w - \nabla_w L$ which can be computed as
\begin{align}
w \gets w \ + \ \delta \ \nabla_w q(s,a;w),  
\end{align}
where $\nabla_w q$ is the gradient of $q$ with respect to the parameters $w$. This process then repeats again, starting at location $s'$ and taking action $a'$ until $s$ is a terminal state (goal) or the number of steps exceeds the maximum steps $T$.

\begin{algorithm*}[hbt!]
			\textbf{Input: } policy $\pi$ to be evaluated \\
			\textbf{Initialize: } $q(s,a;w)$.
			\begin{algorithmic}
				\Repeat{ (for each episode)}
				\State Initialize $s$
				\State Compute $q(s,a;w)$
				\State Choose action $a$ given by extrapolation policy, $\epsilon$-greedy in Eq. \eqref{eq:eps-greedy}
				\Repeat{ (for each step $t$ of episode)}				 
				\State Take action $a$, observe reward $r$ and next state $s'$
				\State Compute $q(s',a';w)$
				\State Choose action $a'$ given by  $\epsilon$-greedy policy in Eq. \eqref{eq:eps-greedy}
				\State Compute the TD error $\delta \gets r + q(s',a';w) - q(s,a;w)$
				\State Update the parameters $w \gets w + \alpha \delta \nabla_w q(s,a;w)$  
				\State $s \gets s'$, $a \gets a'$
				\Until{($s$ is terminal or reaching to max number of steps $T$)}
				\Until{ (convergence or reaching to max number of episodes)}
			\end{algorithmic}
	\caption{SARSA: On-Policy TD Learning}
	\label{Algo:SARSA}
\end{algorithm*}

%% file: sections/sparse.tex
\subsection{Lateral inhibition}

Lateral inhibition can lead to sparse distributed
representations \citep{OReillyRC:2001:CECN} by making a small
and relatively constant fraction of the artificial neurons active at
any one time (e.g., 10\% to 25\%). Such representations achieve a
balance between the generalization benefits of overlapping
representations and the interference avoidance offered by sparse
representations. Another way of viewing the sparse distributed
representations produced by lateral inhibition is in terms of a
balance between competition and cooperation between neurons participating in the representation.

It is important to note that sparsity can be produced in distributed
representations by adding regularization terms to the learning loss
function, providing a penalty during optimization for weights that
cause too many units to be active at
once \citep{French:1991:Sharpenning,Zhang:2015:Sparse-Survay,Liu2018SparseRL}. This learning process is not
necessary, however, when lateral inhibition is used to produce sparse
distributed representations. With this method, feedforward and
feedback inhibition enforce sparsity from the very beginning of the
learning process, offering the benefits of sparse distributed
representations even early in the reinforcement learning process.

\subsection{k-Winners-Take-All ($k$WTA) mechanism}
Computational cognitive neuroscience models have shown that fast
pooled lateral inhibition produces patterns of activation that can be
roughly described as $k$-Winners-Take-All ($k$WTA)
dynamics \citep{OReillyRC:2001:CECN}. A $k$WTA function ensures that
approximately $k$ units out of the $n$ total units in a hidden layer are
strongly active at any given time. Applying a $k$WTA function to the
net input in a hidden layer gives rise to a sparse distributed
representations, and this happens without the
need to solve any form of constrained optimization problem. The $k$WTA function is provided in Algorithm \ref{Algo:kWTA}. The $k$WTA mechanism only requires sorting the net input vector in the hidden layer in every feedforward direction to find the top $k+1$ active neurons. Consequently, it has at most $\textrm{O}(n + k \log k)$ computational time complexity using a partial quicksort algorithm, where $n$ is the number of neurons in the largest hidden layer, and $k$ is the number of winner neurons. $k$ is relatively smaller than $n$. For example $k = 0.1 \times n$ is considered for the simulations reported in this chapter, and this ratio is commonly used in the literature.   
\begin{algorithm*}
	\begin{algorithmic}
		\State Input $\eta$: net input to the hidden layer
		\State Input $k$: number of winner units
		\State Input constant parameter $0<q<1$, e.g., $q= 0.25$
		\State Find top $k+1$ most active neurons by sorting $\eta$, and store them in $\eta'$ in descending order
		\State Compute $k$WTA bias, $b \gets \eta'_k - q (\eta'_k - \eta'_{k+1} )$\\
		\Return $\eta_{kWTA} \gets \eta - b$
	\end{algorithmic}
	\caption{The $k$-Winners-Take-All Function}
	\label{Algo:kWTA}
\end{algorithm*}

\subsection{Feedforward kWTA neural network}
\begin{figure*}[hbt!]
\centering
\includegraphics[width=0.8\textwidth]{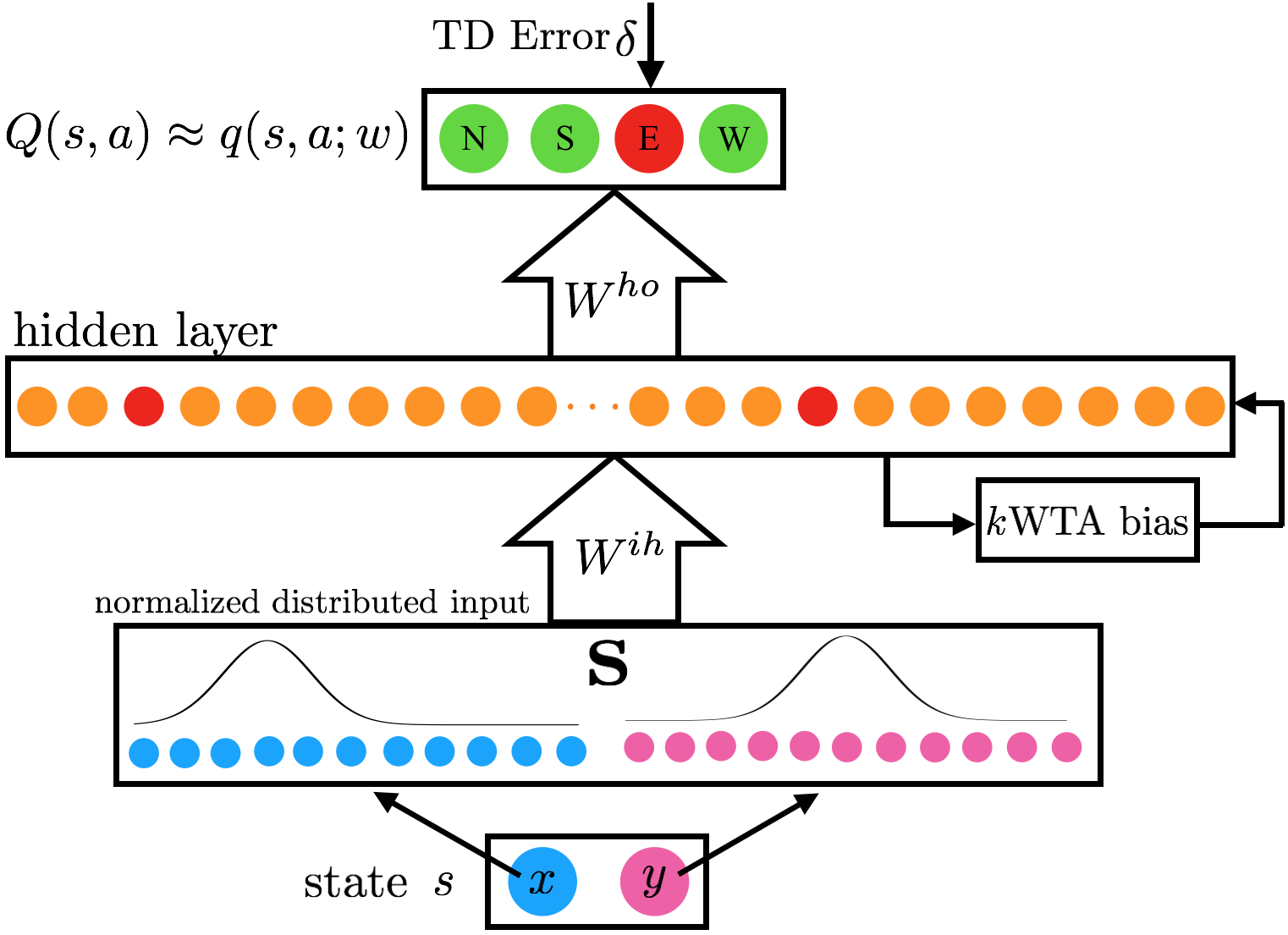}
\caption{The $k$WTA neural network architecture: a backpropagation network with a single layer equipped with the $k$-Winner-Take-All mechanism (from Algorithm \ref{Algo:kWTA}). The $k$WTA bias is subtracted from the hidden units net input that causes polarized activity which supports the sparse conjunctive representation. Only 10\% of the neurons in the hidden layer have high activation. Compare the population of red (winner) neurons to the orange (loser) ones. }
	\label{fig:kwta-network}
\end{figure*}

In order to bias a neural network toward learning sparse conjunctive codes for sensory state inputs, we constructed a variant of a backpropagation neural network architecture with a single hidden layer (see Figure \ref{fig:kwta-network}) that utilizes the $k$WTA mechanism described in Algorithm \ref{Algo:kWTA}.   

Consider a continuous control task (such as Puddle-world in Figure \ref{fig:puddle-world-env}) where the state of the agent is described as 2D coordinates, $s = (x,y)$. Suppose that the agent has to choose between four available actions $\mathcal{A}=$$\{$North, South, East, West$\}$. Suppose that the $x$ coordinate and the $y$ coordinate are in the range $[0,1]$ and each $x$ and $y$ range is discretized uniformly to $n_x$ and $n_y$ points correspondingly. Let's denote $\mathbf{X}=[0:1/n_x:1]$, and $\mathbf{Y}=[0:1/n_y:1]$ as the discretized vectors, i.e., $\mathbf{X}$ is a vector with $n_x+1$ elements from 0 to 1, and all points between them with the grid size $1/n_x$. To encode a coordinate value for input to the network, a Gaussian distribution with a peak value of $1$, a standard deviation of $\sigma_x = 1 / n_x$ for the $x$ coordinate, and $\sigma_y = 1/n_y$ for the $y$ coordinate, and a mean equal to the given continuous coordinate value, $\mu_x=x$, and $\mu_y=y$, was used to calculate the activity of each of the $\mathbf{X}$ and $\mathbf{Y}$ input units. Let's denote the results as $\mathbf{x}$ and $\mathbf{y}$. The input to the network is a concatenation vector $\mathbf{s} := (\mathbf{x},\mathbf{y})$.

We can calculate the \emph{net input} values of hidden units based on the
network inputs, i.e., the weighted sum of the inputs
\begin{align}
\eta:= W^{ih} \mathbf{s} + b^{ih},
\end{align}
where $W^{ih}$ are weights, and $b^{ih}$ are biases from the input layer to the hidden layer. After calculating the net input, we compute the $k$WTA bias, $b$, using Algorithm \ref{Algo:kWTA}. We subtract $b$ from all of the net input values, $\eta$, so that the $k$ hidden units with the highest net input values have positive net input values, while all of the other hidden units adjusted net input values become negative. These adjusted net input values, i.e. $\eta_{kwta} = \eta - b$ were transformed into unit activation values using a logistic sigmoid activation function (gain of $1$, offset of $-1$), resulting in hidden unit activation values in the range between $0.0$ and $1.0$, 
\begin{align}
\mathbf{h} = \frac{1}{1 + e^{-(\eta_{kwta} - 1)}}
\end{align}
with the  top $k$ units having activations above $0.27$ (due to the $-1$ offset), and the ``losing'' hidden units having activations below that
value. The $k$ parameter controlled the degree of sparseness of the
hidden layer activation patterns, with low values producing more
sparsity (i.e., fewer hidden units with high activations). In the
simulations of this paper, we set $k$ to be $10\%$ of the total number
of hidden units. The output layer of the $k$WTA neural network is fully connected to the hidden layer and has $|\mathcal{A}|$ units, with each unit $i=1,\dots,|\mathcal{A}|$ representing the state-action values $q(s,a_i;w)$. We calculate these values by computing the activation in the output layer
\begin{align}
\mathbf{q} = W^{ho} \mathbf{h} + b^{ho},
\end{align}
where $W^{ho}$ are weights, and $b^{ho}$ are biases, from the hidden layer to the output layer. The output units used a linear activation function, hence, $\mathbf{q}$ is a vector of the state-action values.

In addition to encouraging sparse distributed representations, this
$k$WTA mechanism has two properties that are worthy of note. First,
introducing this highly nonlinear mechanism violates some of the
assumptions relating the backpropagation of error procedure to
stochastic gradient descent in error. Thus, the connection weight
changes recommended by the backpropagation procedure may slightly
deviate from those which would lead to local error minimization in
this network. We opted to ignore this discrepancy, however, trusting
that a sufficiently small learning rate would keep these deviations
small. Second, it is worth noting that this particular $k$WTA
mechanism  allows for a distributed pattern of activity over the hidden
units, making use of intermediate levels of activation. This provides
the learning algorithm with some flexibility, allowing for a graded
range of activation levels when doing so reduces network error. As
connection weights from the inputs to the hidden units grow in
magnitude, however, this mechanism will drive the activation of the
top $k$ hidden units closer to $1$ and the others closer to
$0$. Indeed, an examination of the hidden layer activation patterns in
the $k$WTA-equipped networks used in this study revealed that the $k$
winning units consistently had activity levels close to the maximum
possible value, once the learning process was complete.

%% file: sections/simulation-tasks.tex
\subsection{Numerical simulations design}
\begin{figure*}
	\begin{center}
		\begin{tabular}{ccc}
			\includegraphics[width=0.28\textwidth]{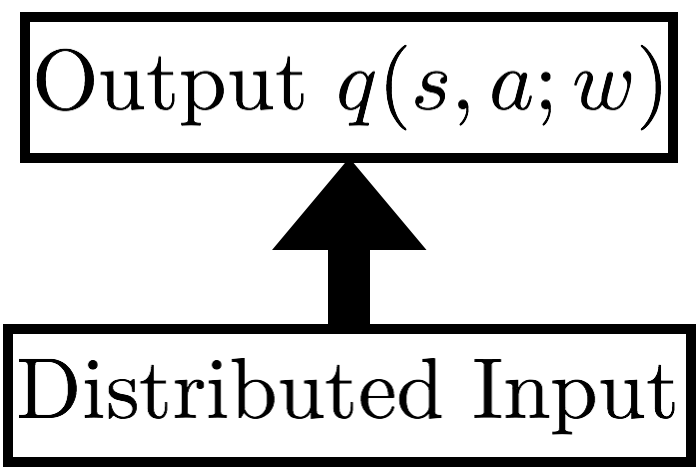}   &
			\includegraphics[width=0.28\textwidth]{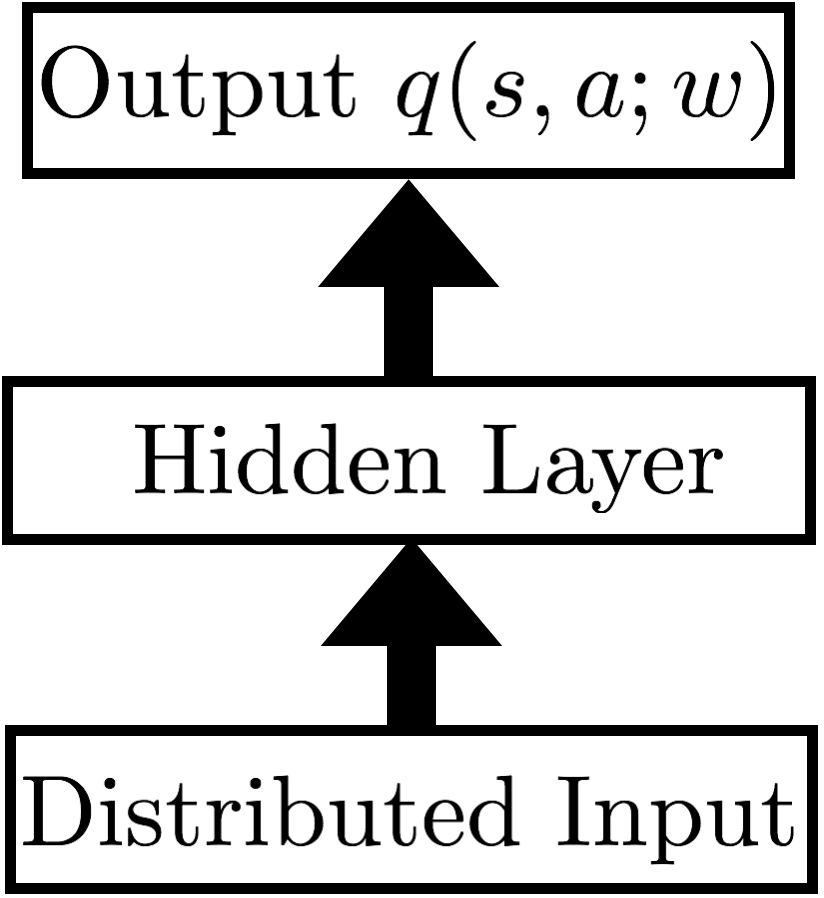}       &
			\includegraphics[width=0.28\textwidth]{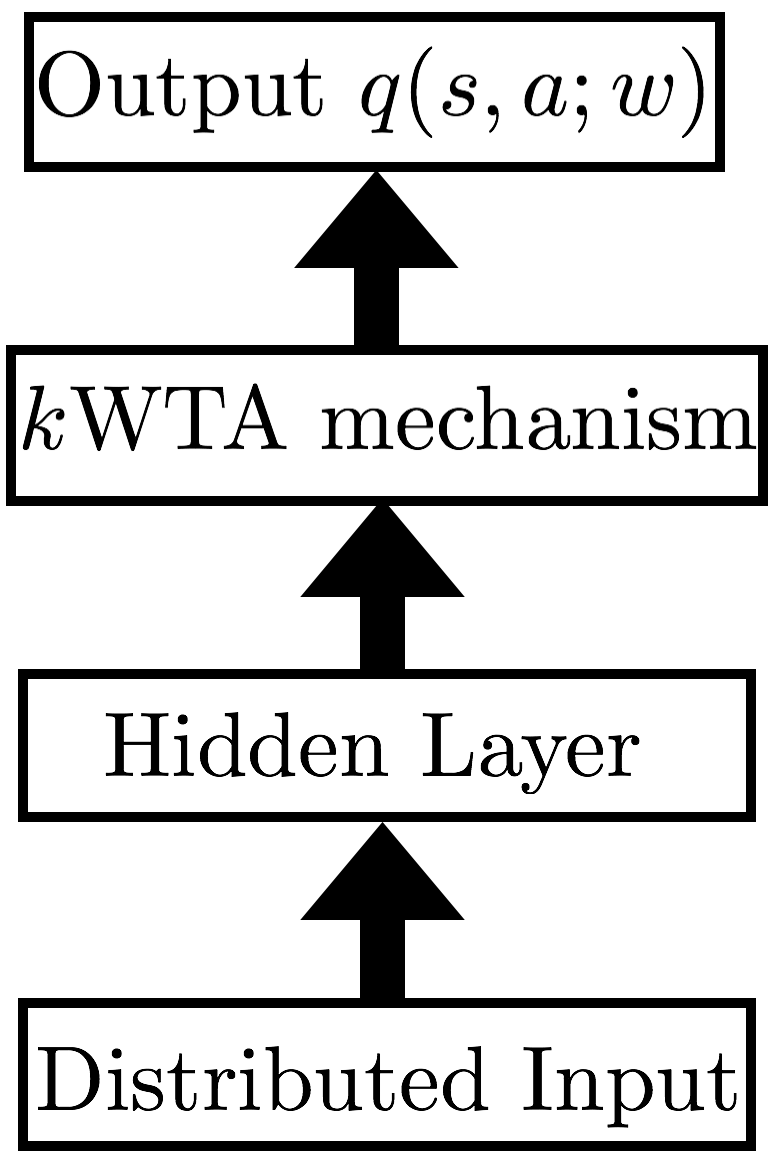} \\
				{(a) Linear}  & {(b) Regular} & {(c) $k$WTA}
		\end{tabular}
	\end{center}
	\caption{The neural network architectures used as the function approximator for state action values $q(s,a;w)$. (a) Linear network. (b) Regular backpropagation neural network. (c) $k$WTA network.}
	\label{fig:networks}
\end{figure*}

In order to assess our hypothesis that biasing a neural network toward
learning sparse conjunctive codes for sensory state inputs will
improve TD Learning when using a neural network as a function approximator for $q(s,a;w)$ state-action value function, we constructed three types of backpropagation networks: 
\begin{description}
	\item[kWTA network.] A single layer backpropagation neural network equipped with the $k$-Winners-Take-All mechanism. See Figures \ref{fig:kwta-network} and \ref{fig:networks}(c). 
	\item[Regular network.] A single layer backpropagation neural network without the $k$WTA mechanism. See Figure \ref{fig:networks}(b).
	\item[Linear network.] A linear neural network without a hidden layer. See Figure \ref{fig:networks}(a)
\end{description}

There was complete connectivity between the input units
and the hidden units and between the hidden units and the output
units. For Linear networks, there was full connectivity between the input layer and the output layer. All connection weights were initialized to uniformly sampled random values in the range $[-0.05,0.05]$.

In order to investigate the utility of sparse distributed
representations, simulations were conducted involving three relatively
simple reinforcement learning control tasks: the \emph{Puddle-world}
task, the \emph{Mountain-car} task, and the
\emph{Acrobot} task. These reinforcement learning problems were
selected because of their extensive use in the literature \citep{Boyan:1995:Approximating,SuttonRS:1996:Coarse}. In this
section, each task is described. We tested the SARSA variant
of TD Learning~\citep{Sutton:Barto:2017:RL-BOOK-2}. (See
Algorithm~\ref{Algo:SARSA}.) on each of the three neural networks architectures. The Matlab code for these simulations is available at \texttt{http://rafati.net/td-sparse/}.

The specific parameters for each simulation can be found in the description of the simulation tasks below. In Section ``Results and Discussions'', the numerical results for training performance for each task are reported and discussed. 

\subsection{The Puddle-world task}
\begin{figure}[hbt!]
	\centering
	\includegraphics[width=0.8\textwidth]{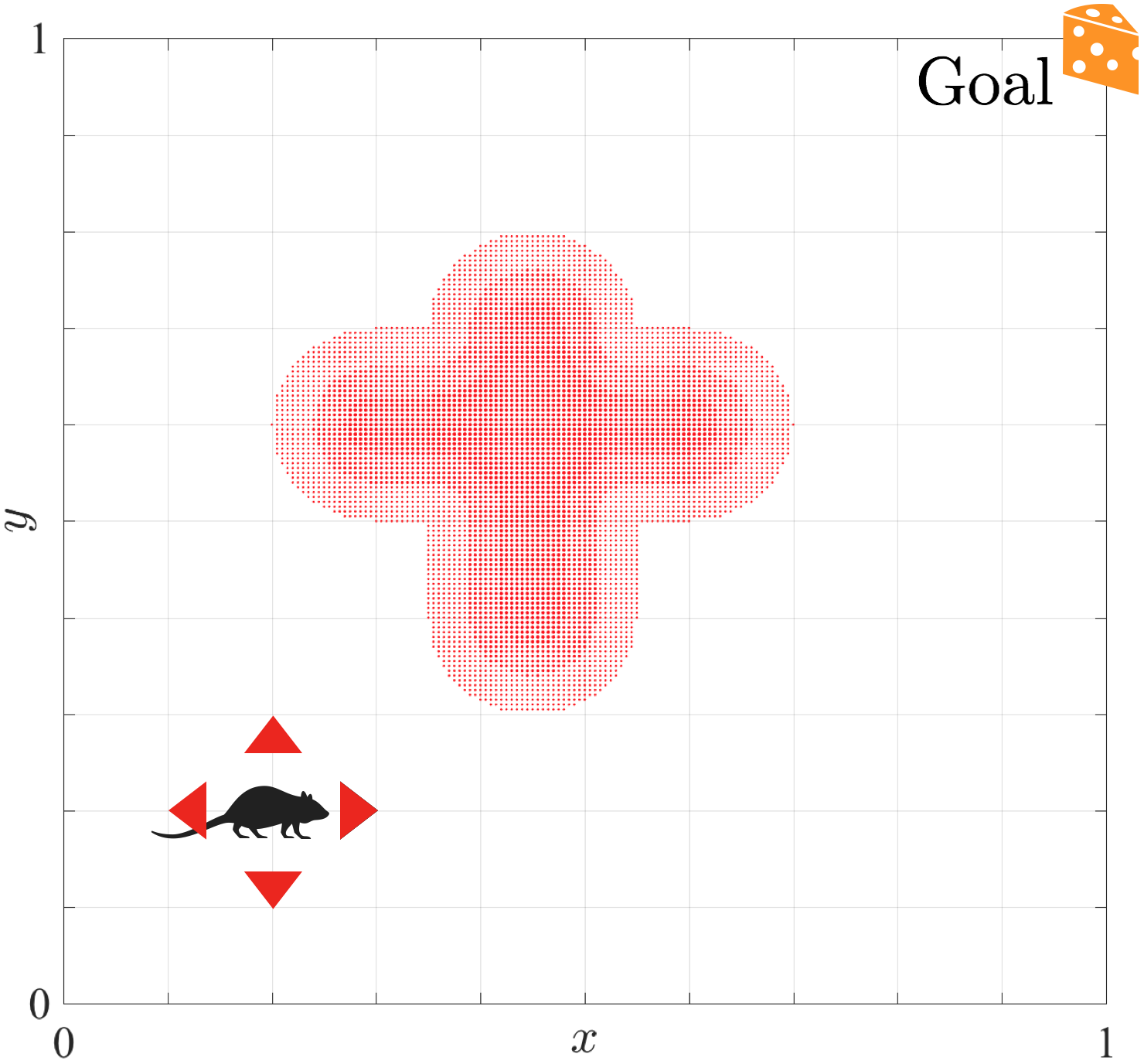}
	\caption{The agent in puddle-world task attempts to reach the goal location (fixed in the Northeast corner) in the least time steps by avoiding the puddle. The agent moves a distance of $0.05$ either North, South, East, or West on each time step. Entering a puddle produces a reward of $(-400 \times d)$, where $d$ is the distance of the current location to the edge of the puddle. This value was $-1$ for most of the environment, but it had a higher value, $0$, at the goal location in the Northeast corner. Finally, the agent receives a reward signal of $-2$ if it had just attempted to leave the square environment. This pattern of reinforcement was selected to parallel that previously used in \cite{SuttonRS:1996:Coarse}.}
	\label{fig:puddle-world-env}
\end{figure}

The agent in the Puddle-world task attempts to navigate in a dark 2D grid world to reach a goal location at the top right corner, and it should avoid entering poisonous ``puddle'' regions (Figure~\ref{fig:puddle-world-env}). In every episode of Algorithm \ref{Algo:SARSA}, the agent is located in a random state. The agent can choose to move in one of four directions, $\mathcal{A} = \{$North, South, East, West$\}$. For the Puddle-world task, the $x$-coordinate and the $y$-coordinate of the current state, $s$, were presented to the neural network over two separate pools of input units. Note that these coordinate values were in the range $[0,1]$, as shown in Figure \ref{fig:puddle-world-env}. 
Each pool of input units consisted of $21$ units, with each unit corresponding to a coordinate location between $0$ and $1$, inclusive, in increments of $0.05$. To encode a coordinate value for input to the network, a Gaussian distribution with a peak value of $1$, a standard deviation of $0.05$, and a mean equal to the given continuous coordinate value was used to calculate the activity of each of the $21$ input units (see Figure \ref{fig:kwta-network}).  

We use each of the three mentioned neural network models as function approximators for state-action values (see Figure \ref{fig:networks}). All networks had four output units, with each output corresponding to one of the
four directions of motion. The hidden layer for both regular BP and $k$WTA neural networks had $220$ hidden units. In the $k$WTA network, only $10\%$ (or $22$) of the hidden units were allowed to be highly active. 

At the beginning of the simulation, the exploration probability, $\epsilon$, was set to a relatively high value of $0.1$, and it remained at this value for much of the learning process. Once the average magnitude of $\delta$ over an episode fell below $0.2$, the value of $\epsilon$ was reduced by $0.1\%$ each time the goal location was reached. Thus, as the agent became increasingly successful at reaching the goal location, the
exploration probability, $\epsilon$, approached zero. (Annealing the
exploration probability is commonly done in systems using TD
Learning.) The agent continued to explore the environment, one episode
after another, until the average absolute value of $\delta$ was below
$0.01$ and the goal location was consistently reached, or a maximum of
$44,100$ episodes had been completed. This value was heuristically
selected as a function of the size of the environment: $(21 \times 21)
\times 100 = 44,100$. Each episode of SARSA was terminated if the agent had reached the goal in the corner of the grid or after the maximum steps, $T = 80$, had been taken. The learning rate remained fixed $\alpha=0.005$ during training.

When this reinforcement learning process was complete, we examined
both the behavior of the agent and the degree to which its value
function approximations, $q(s,a;w)$, matched the correct values determined by running SARSA to convergence while using a large look-up table to capture the value
function.

\subsection{The Mountain-car task}
\begin{figure}[hbt!]
\centering
\includegraphics[width=0.8\textwidth]{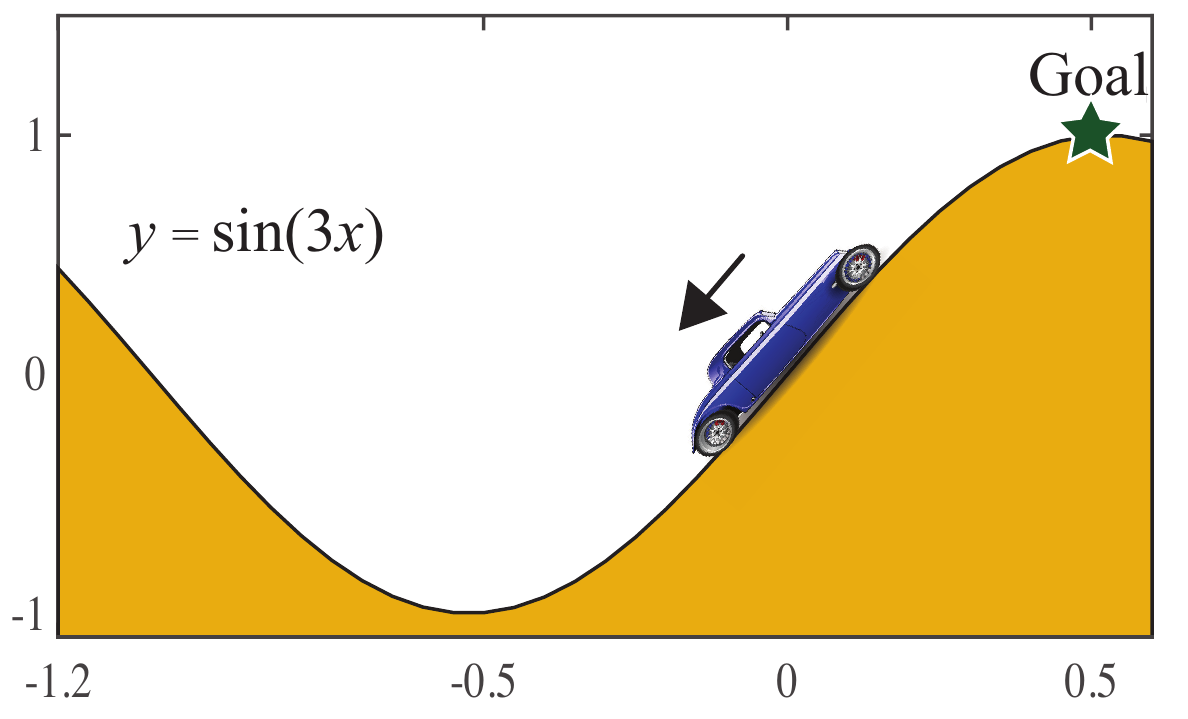}
\caption{The goal is to drive an underpowered car up a steep
		hill. The agent received -1 reward for each time step until it
		reached the goal, at which point it received 0 reward.}
\label{fig:mountain_car}
\end{figure}

In this reinforcement learning problem, the task involves driving a car up a steep mountain road to a high goal location. The task is difficult
because the force of gravity is stronger than the car's engine (see Figure \ref{fig:mountain_car}). In order to solve the problem, the agent must learn first to move away from the goal, then use the stored potential energy in combination with the engine to overcome gravity and reach the goal state.
The state of the Mountain-car agent is described by the car's position and
velocity, $s=(x,\dot{x})$. There are three possible actions: $\mathcal{A}=$\texttt{\{forward, neutral, backward\}} throttle of the
motor. After choosing action $a \in \mathcal{A}$, the car's state is
updated by the following equations
\begin{subequations}
\begin{align}
x_{t+1} &= \text{bound} (x_t + \dot{x}_{t+1}) \\
\dot{x}_{t+1} &= \text{bound} (\dot{x}+ 0.0001 a_t - 0.0025 \cos(3 x_t)),
\end{align}
\end{subequations}
where the bound function keeps state variables within their limits, $x
\in [-1.2,0.5]$ and $\dot{x} \in [-0.07,0.07]$. If the car reaches
the left environment boundary, its velocity is reset to zero. In order to present the values of the state variables to the neural network, each variable was encoded as a Gaussian activation bump surrounding the input
corresponding to the variable value. Every episode started from a
random position and velocity. Both the $x$-coordinate
and the $\dot{x}$ velocity was discretized to 60 mesh points, allowing each variable to be represented over 61 inputs. To encode a state value for input to the network, a Gaussian distribution was used to calculate the
activity of each of the $61$ input units for that variable. The
network had a total of $122$ inputs.

All networks had three output units, each corresponding to one of the
possible actions. Between the $122$ input units and the $3$ output units was a layer of $61 \times 61 \times 0.7 = 2604$ hidden units (for $k$WTA and Regular networks). The hidden layer of the $k$WTA network was subject to the previously described $k$WTA mechanism, parameterized so as to allow $10\%$, or $260$, of the hidden units to be highly active.  

We used the SARSA version of TD Learning
(Algorithm~\ref{Algo:SARSA}). The exploration probability, $\epsilon$,
was initialized to $0.1$ and it was decreased by $1\%$ after each
episode until reaching a value of $0.0001$. The agent received a
reward of $r=-1$ for most of the states, but it received a reward of
$r=0$ at the goal location ($x \geq 0.5$). If the car collided with
the leftmost boundary, the velocity was reset to zero, but there was
no extra punishment for bumping into the wall. The learning rate, $\alpha=0.001$, stayed fixed during the learning.  
The agent explored the Mountain-car environment in
\emph{episodes}. Each episode began with the agent being placed at a
location within the environment, sampled uniformly at random. Actions
were then taken, and connection weights updated, as described
above. The episode ended when the agent reached the goal location or
after the maximum of $T = 3000$ actions had been taken. The agent continued to explore the environment, one episode
after another, until the average absolute value of $\delta$ was below
$0.05$ and the goal location was consistently reached, or a maximum of
$200,000$ episodes had been completed.

\subsection{The Acrobot task}
\begin{figure}[hbt!]
\centering
	\includegraphics[width=0.7\textwidth]{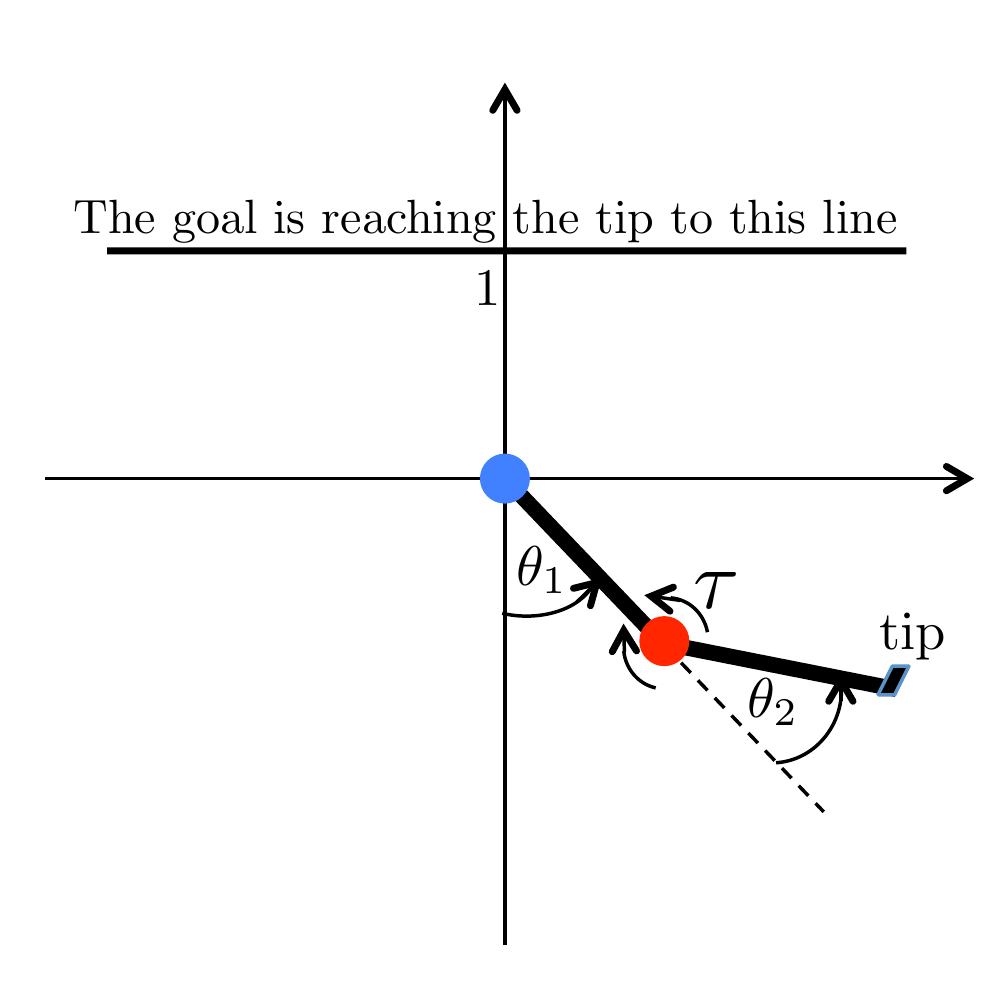}
	\caption{The goal is to swing the tip (``feet'') above the horizontal by the length of the lower ``leg'' link. The agent receives -1 reward until it reaches to goal, at which point it receives $0$ reward.}
	\label{fig:acrobot}
\end{figure}

We also examined the utility of learning sparse distributed representations on the Acrobot control task, which is a more complicated task and one attempted by~\cite{SuttonRS:1996:Coarse}. The \textit{acrobot} is a two-link under-actuated robot, a simple model of a gymnast swinging on a high-bar~\citep{SuttonRS:1996:Coarse}. (See Figure~\ref{fig:acrobot}.) The state of the acrobot is determined by four continuous state variables: two joint angles $(\theta_1,\theta_2)$ and corresponding velocities. Thus, the state the agent can be formally described as $s = (\theta_1,\theta_2,\dot{\theta_1},\dot{\theta_2})$. The goal is to control the acrobot so as to swing the end tip (``feet'') above the horizontal by the length of the lower ``leg'' link. Torque may only be applied to the second joint. The agent receives $-1$ reward until it reaches the goal, at which point it receives a reward of $0$. The frequency of action selection is set to 5 Hz, and the time step, $\Delta t = 0.05 s$, is used for numerical integration of the equations describing the dynamics of the system. A discount factor of $\gamma = 0.99$ is used.
The equations of motion for acrobot are,
\begin{subequations}
	\begin{align}
	\ddot{\theta_1} &= -d_1^{-1} (d_2 \ddot{\theta_2} + \phi_2),\\
	\ddot{\theta_2} &= - \left( m_2 l^{2}_{c2} + I_2 - \frac{d^{2}_2}{d_1} \right)^{-1} \left( \tau + \frac{d_2}{d_1} \phi_1 - \phi_2 \right), 
	\end{align}
\end{subequations}
where $d_1$, $d_2$, $\phi_1$ and $\phi_2$ are defined as
\begin{subequations}
	\begin{align}
	d_1 &= m_1 l^{2}_{c1} + m_2 l_1 l_{c2} \cos(\theta_2) + I_2,  \\
	d_2 &= m_2 \left( l_{c2}^{2} + l_1 l_{c2} \cos(\theta_2) \right) + I_2,\\ 
\begin{split}
	\phi_1 &= - m_2 l_1 l_{c2} \dot{\theta_1}^2 \sin(\theta_2) - 2 m_2 l_1 l_{c2} \dot{\theta_2} \dot{\theta_1}\sin(\theta_2) \\
	&\quad  \quad \quad + (m_1 l_{c1} + m_2 l_{1})g \cos(\theta_1 - \pi / 2) + \phi_2,
\end{split}
  \\
	\phi_2 &= m_2 l_{c2}g \cos(\theta_1 + \theta_2 - \pi / 2).
	\end{align}
\end{subequations}
The agent chooses between three different torque values, $\tau \in \{-1,0,1\}$, with torque only applied to the second joint. The angular velocities are bounded to $\theta_1 \in [-4\pi,4\pi]$ and $\theta_2 \in [-9 \pi, 9\pi]$. The values $m_1=1$ and $m_2=1$ are the masses of the links, and $l_1 = 1 m$ and $l_2 = 1 m$ are the lengths of the links. The values $l_{c1}=l_{c1}=0.5 m$ specify the location of the center of mass of links, and $I_1 = I_2 = 1 kg\ m^2$ are the moments of inertia for the links. Finally, $g = 9.8 m/s^2$ is the acceleration due to gravity. These physical parameters are previously used in \cite{SuttonRS:1996:Coarse}.

In our simulation, each of the four state dimensions is divided over
20 uniformly spaced ranges. To encode a state value for input to the
network, a Gaussian distribution with a peak value of $1$, a standard
deviation of $1/20$, and a mean equal to the given continuous state
variable value was used to calculate the activity of each of the $21$
inputs for that variable. Thus, the network had $84$ total inputs.

The network had three output units, each corresponding to one of the
possible values of torque applied: clockwise, neutral, or
counter-clockwise. Between the $84$ inputs and the $3$ output units
was a layer of $8400$ hidden units for the $k$WTA network and the regular backpropagation network. For the $k$WTA network, only $10\%$, or $840$, of the hidden units were allowed to be highly active. 

The acrobot agent explores its environment in
\emph{episodes}. Each episode of learning starts with the acrobot agent hanging straight down and at rest (i.e., $s=(0,0,0,0)$). The episode ends when the agent reaches the goal location or
after the maximum of $T = 2000$ actions are taken. At the beginning of
a simulation, the exploration probability, $\epsilon$, is set to a
relatively high value of $0.05$. When the agent first reaches the goal, the value of $\epsilon$ starts to decrease, being reduced by $0.1\%$ each time the goal location is reached. Thus, as the agent
becomes increasingly successful at reaching the goal location, the
exploration probability, $\epsilon$, approaches to lower bound of $0.0001$. The agent continues to explore the environment, one episode
after another, until the average absolute value of $\delta$ is below
$0.05$ and the goal location is consistently reached, or a maximum of
$200,000$ episodes are completed. A small learning rate $\alpha=0.0001$ was used and stayed fixed during the learning.

%% file: sections/results.tex
We compared the performance of our $k$WTA neural network with that
produced by using a standard backpropagation network (Regular) with identical parameters. We also examined the performance of a linear network (Linear), which had no hidden units but only complete connections from all input units directly to the four output units (see Figure \ref{fig:networks}).
\subsection{The puddle-world task}
\begin{figure*}[hbt!]
\centering
\begin{tabular}{lccc}
	& \raisebox{5mm}{Linear}  & \raisebox{5mm}{Regular} &  \raisebox{5mm}{$k$WTA} \\ 
	\rotatebox{90}{\hspace{0.5cm}Values} &
	\includegraphics[width=0.28\textwidth]{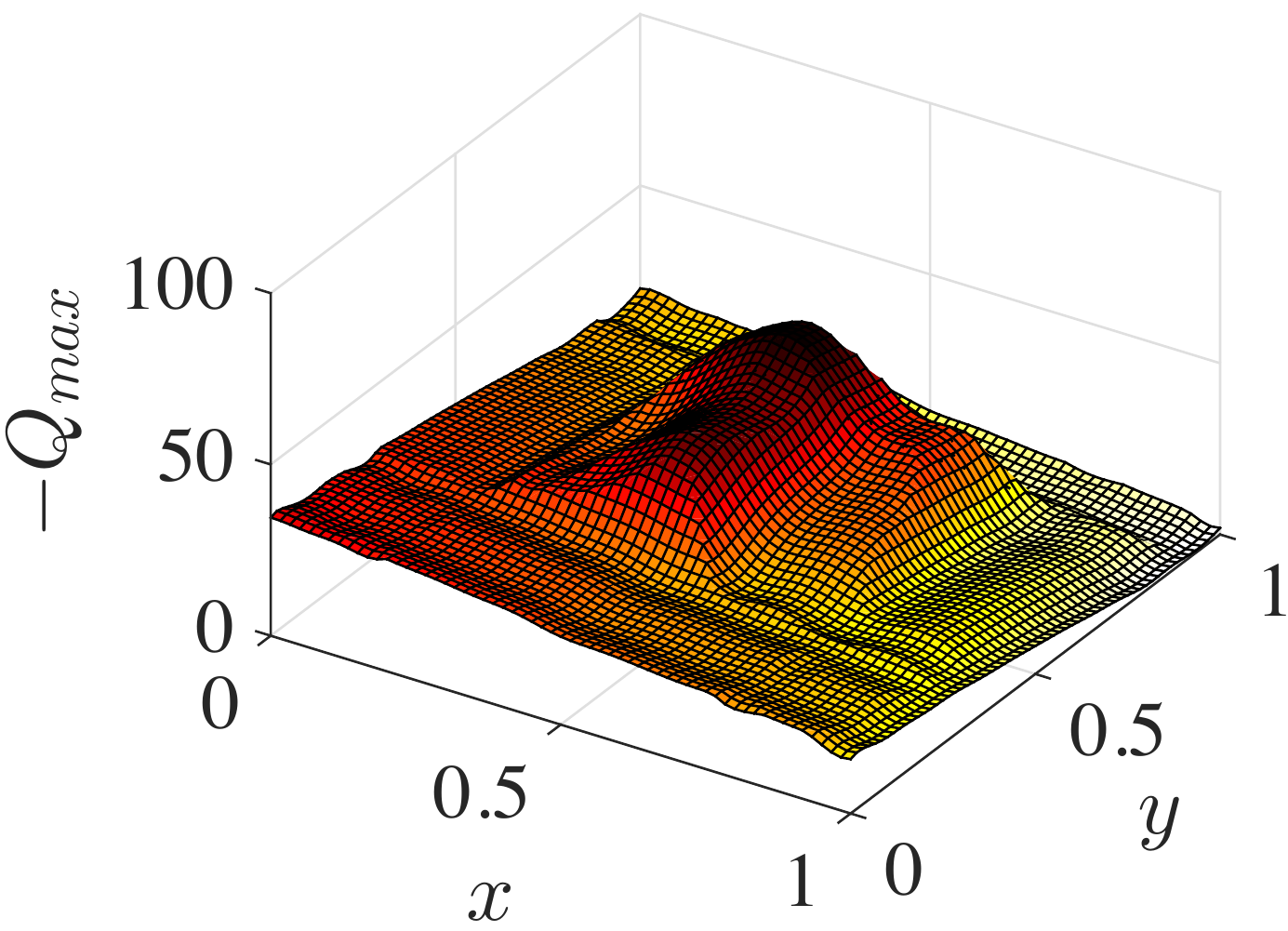}   &
	\includegraphics[width=0.28\textwidth]{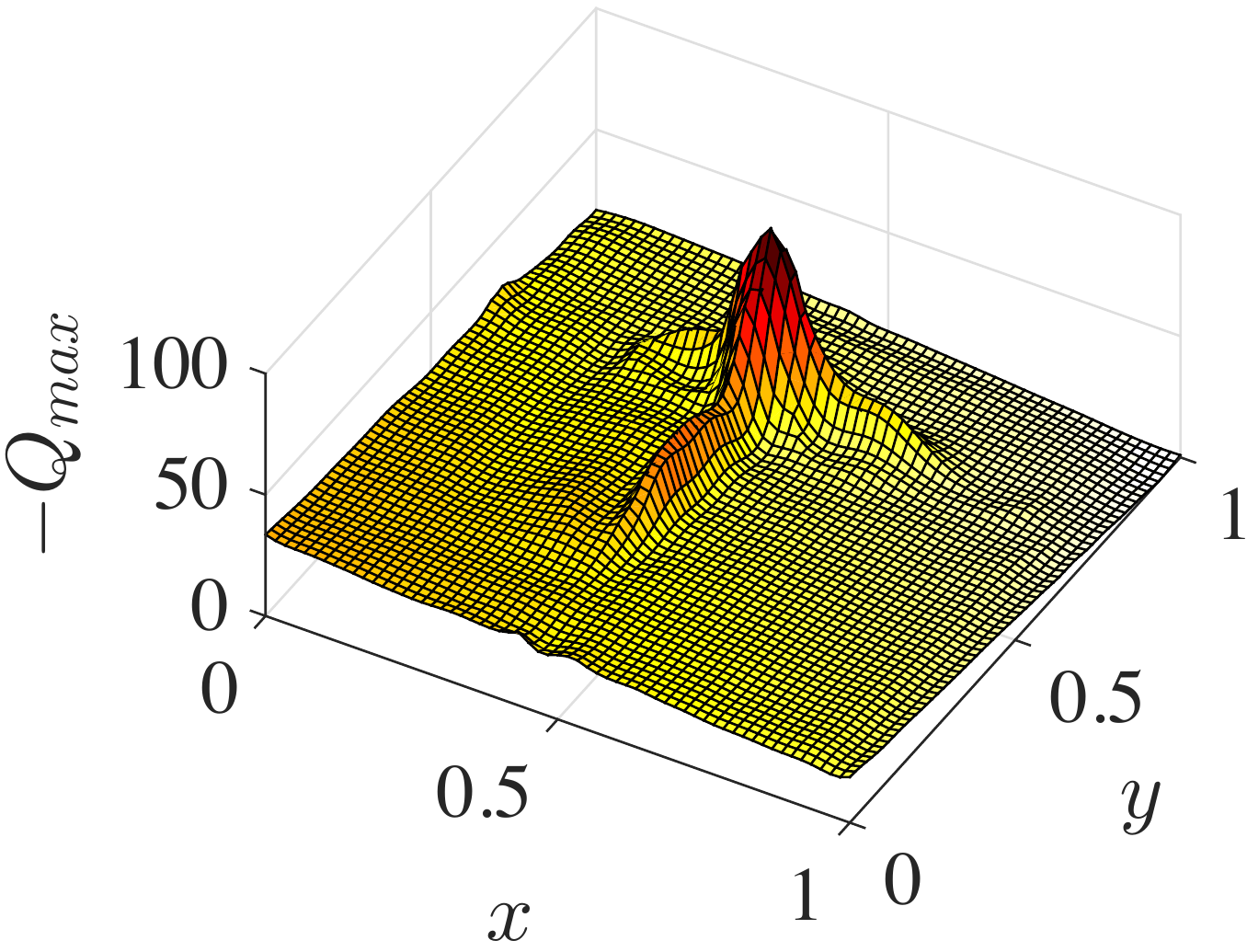}       &
	\includegraphics[width=0.28\textwidth]{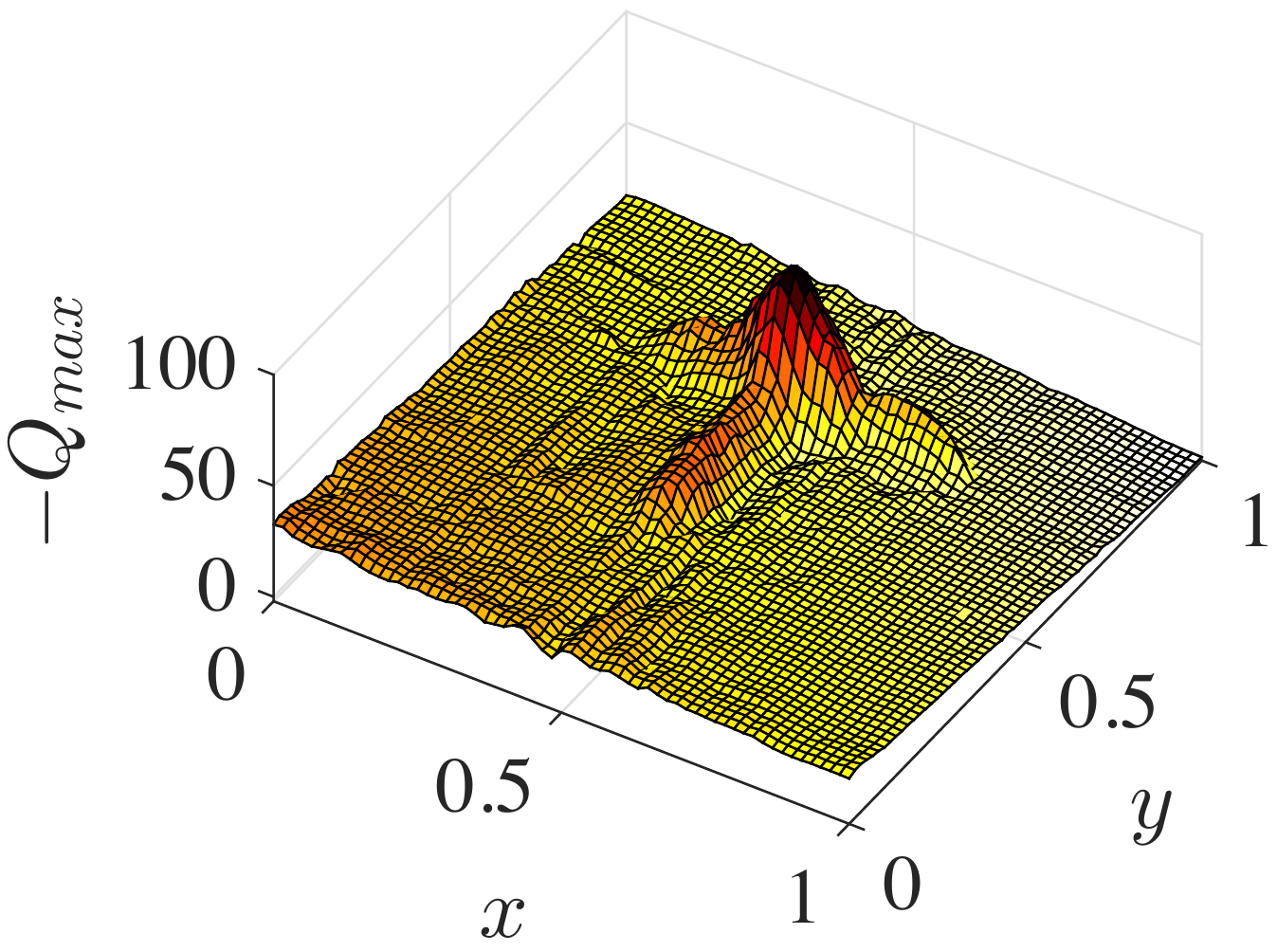} \\
	& (a) & (b) & (c) \vspace{0.1cm}\\
	\rotatebox{90}{\hspace{1.2cm}Policy} &
	\includegraphics[width=0.28\textwidth]{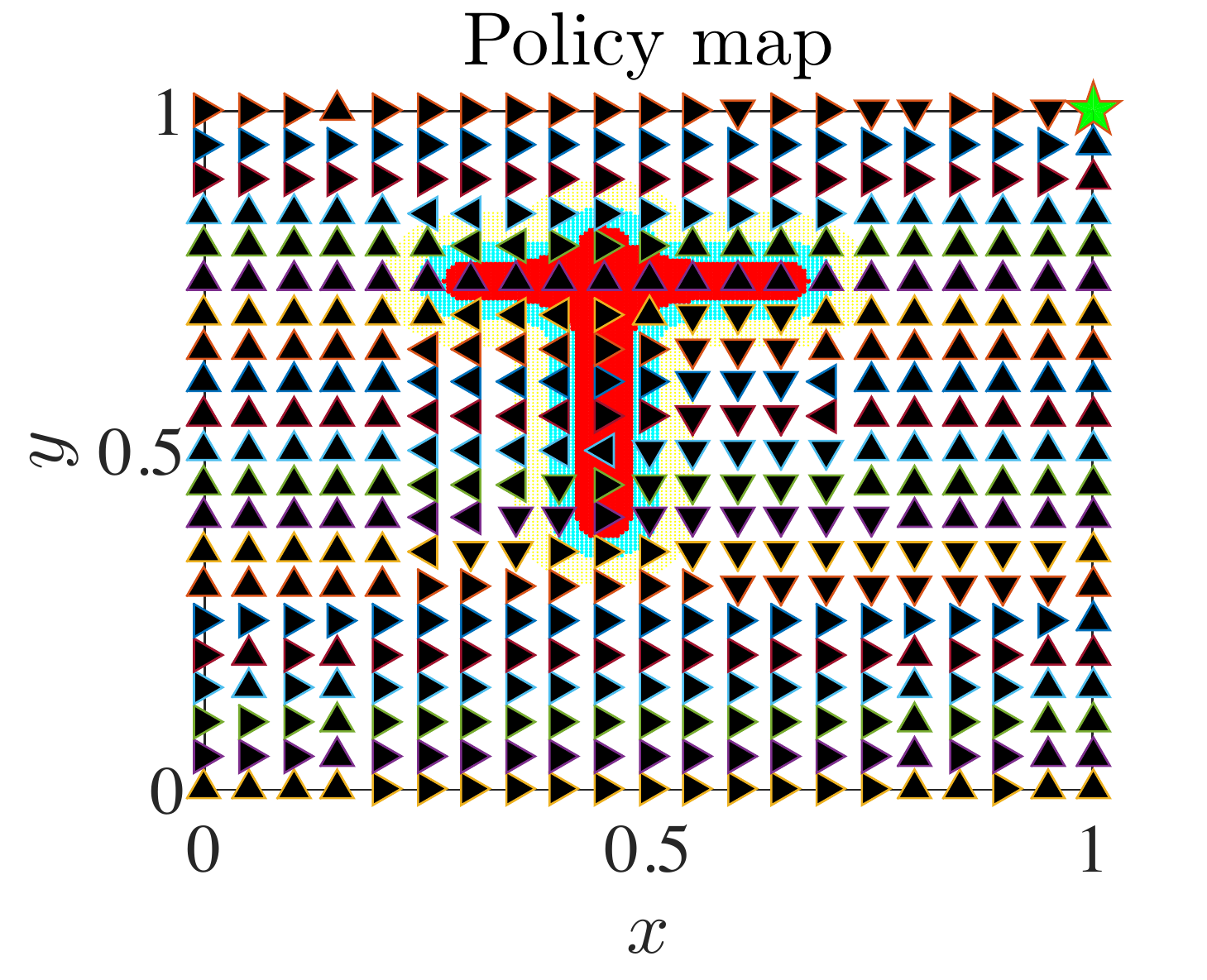}  &
	\includegraphics[width=0.27\textwidth]{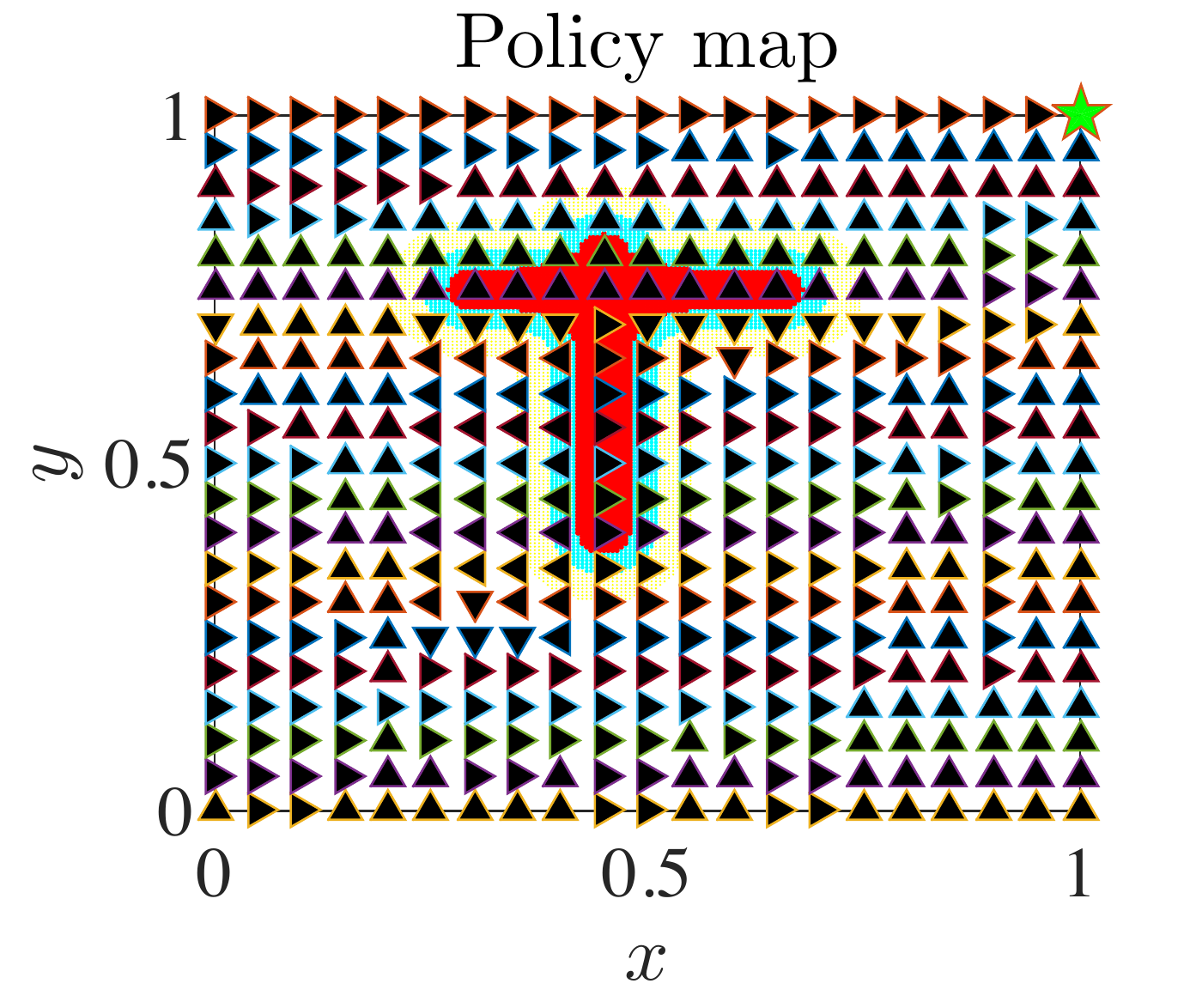}      &
	\includegraphics[width=0.28\textwidth]{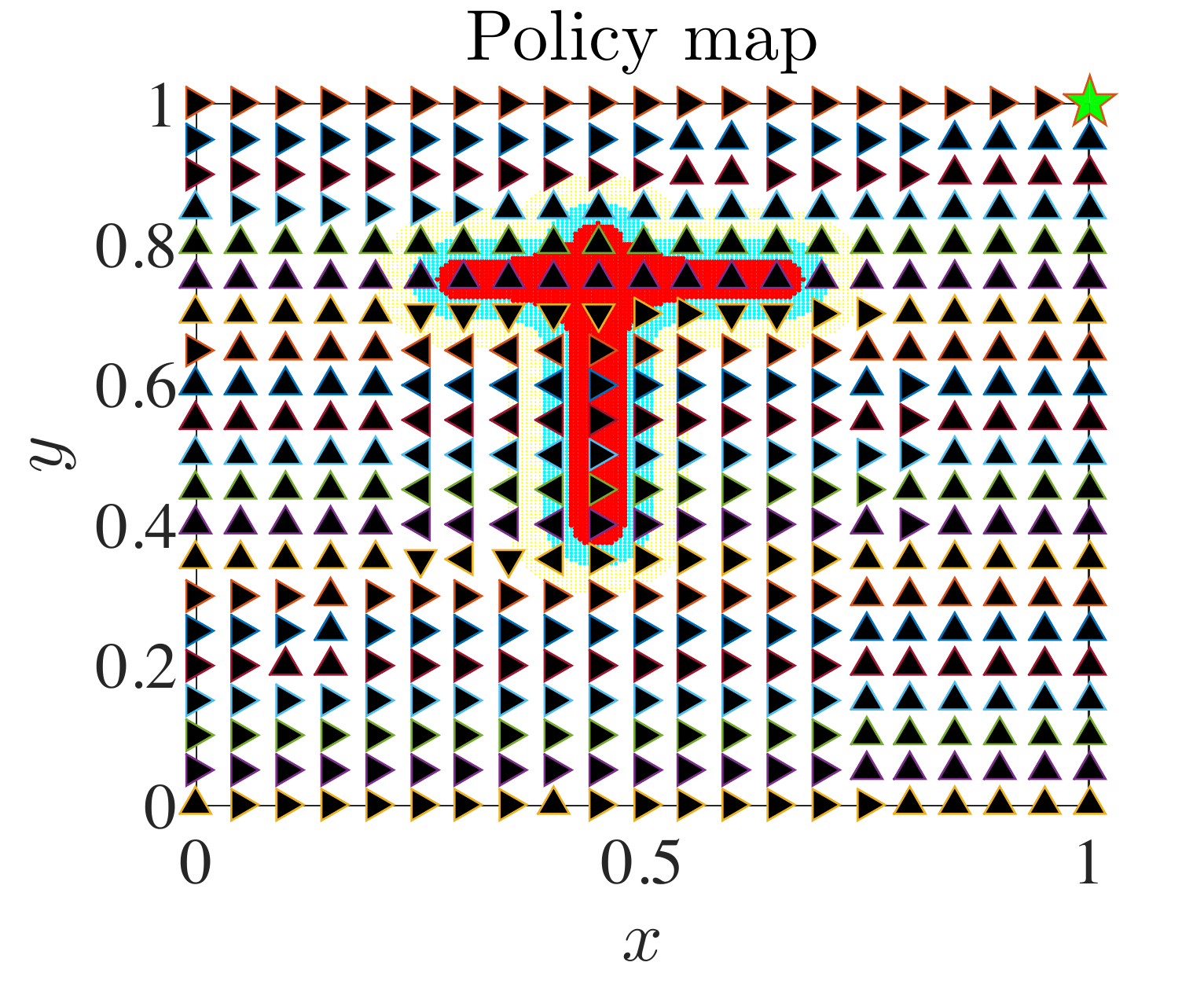} \\
	& (d) & (e) & (f)   \vspace{0.1cm}\\
	\rotatebox{90}{\hspace{1cm}TD Error} &
	\includegraphics[width=0.28\textwidth]{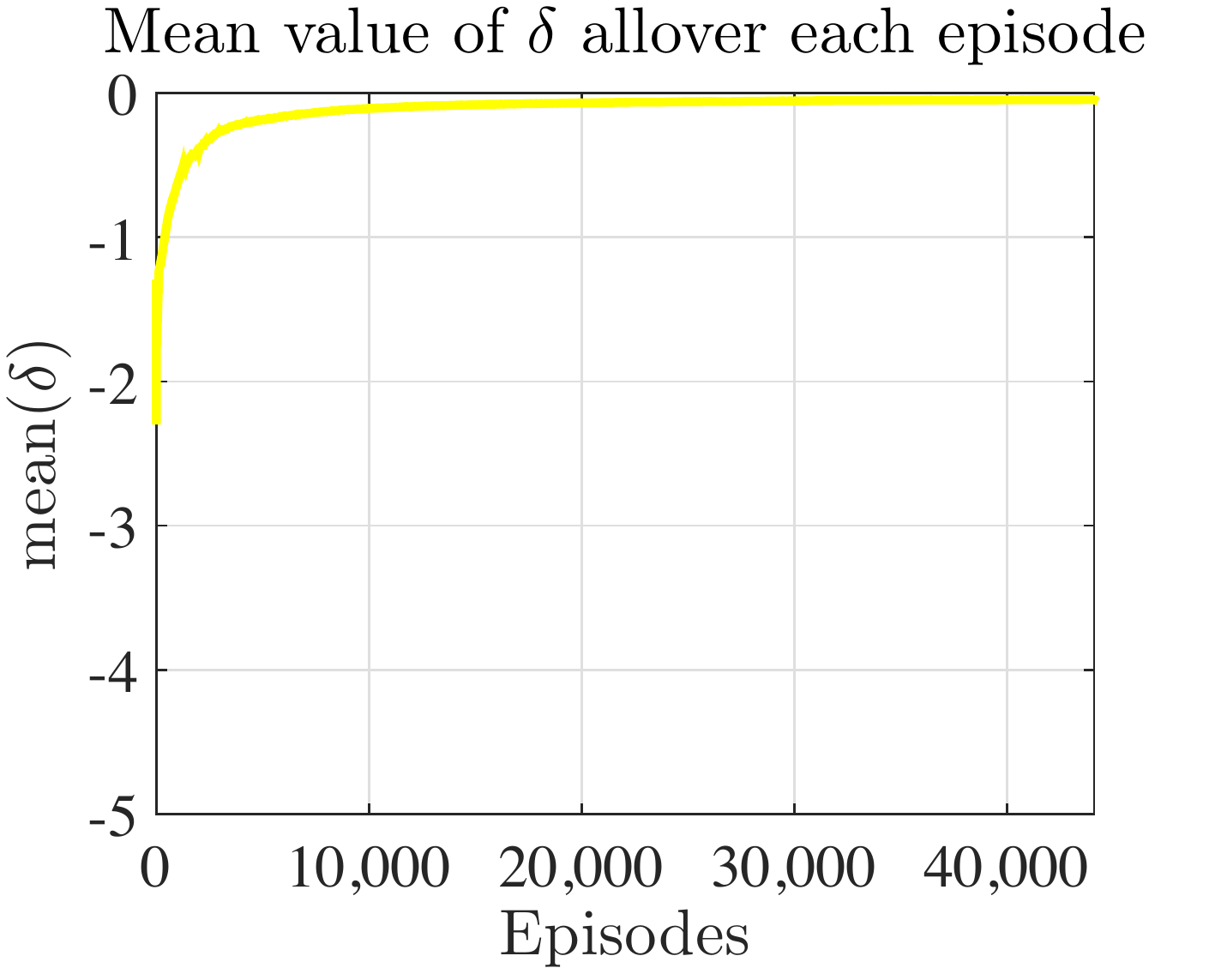}   &
	\includegraphics[width=0.27\textwidth]{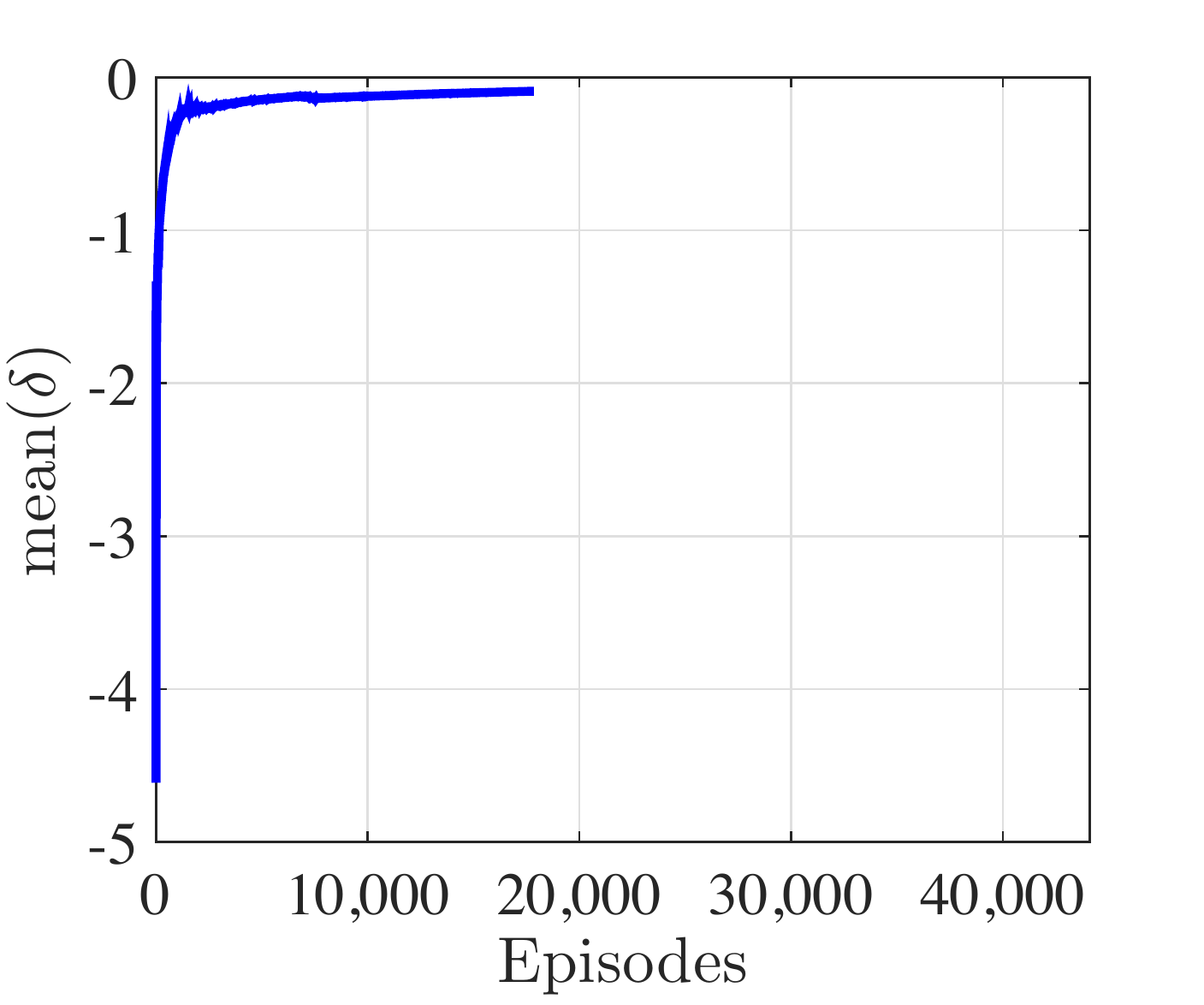}       &
	\includegraphics[width=0.28\textwidth]{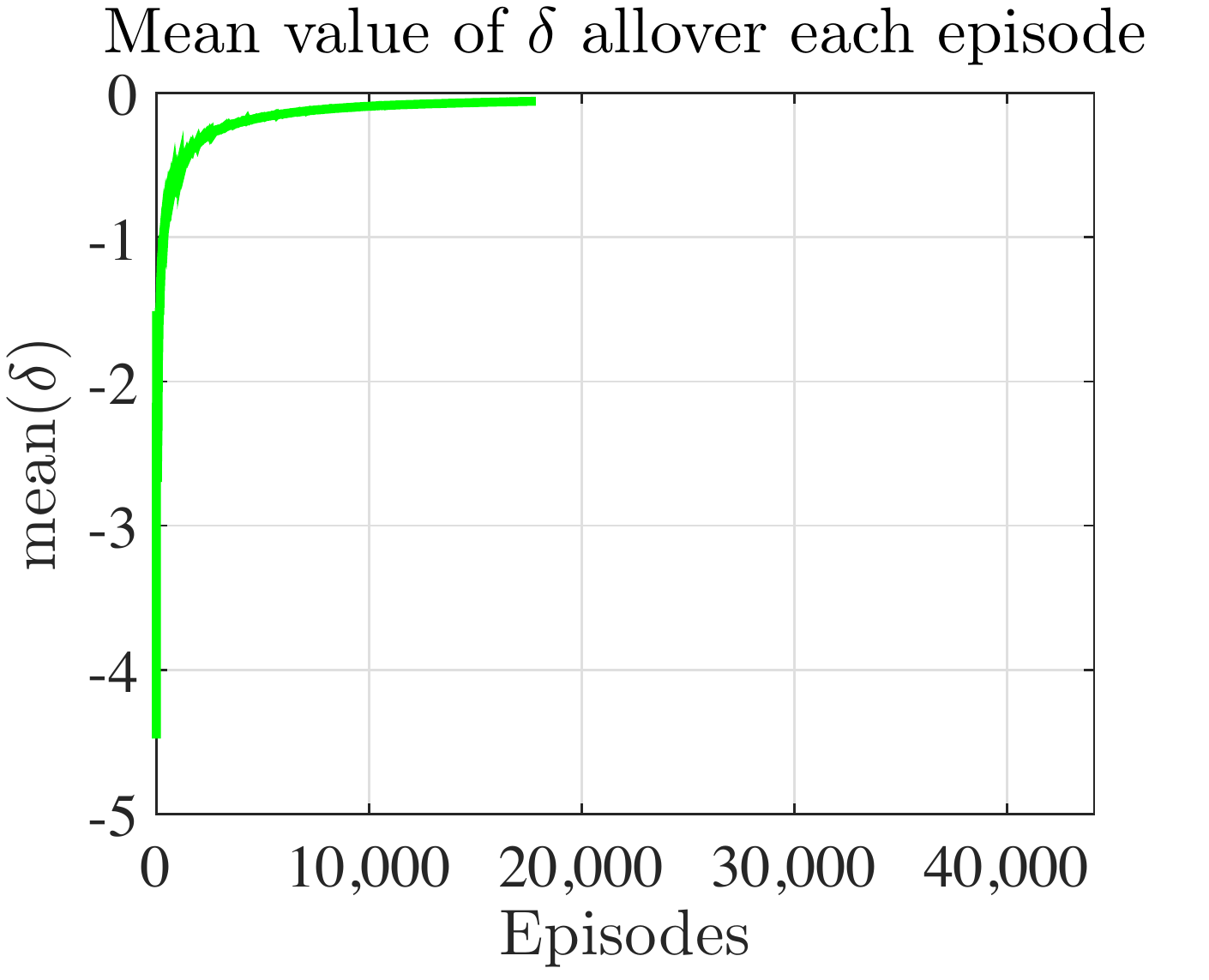}\\
	& (g) & (h) & (i)
\end{tabular}
\caption{The performance of various learned value function approximators may be compared in terms of their success at learning the true value function, the resulting action selection policy, and the amount of experience in the environment needed to learn. The approximate of the state values, expressed as $\mbox{\it max}_{a}\ {Q}(s,a)$ for each state, $s$ is given. (a) Values of states for the Linear network. (b) Values of states for the Regular network. (c) Values of state for $k$WTA network. The actions selected at a grid of locations is shown in the middle row. (d) Policy of states derived from Linear network. (e) Policy of states derived from Regular network. (f) Policy of states derived from $k$WTA network. The learning curve, showing the TD Error over learning episodes, is shown on the bottom. (g) Average TD error for training Linear network. (h) Average TD error for training Regular network. (i) Average TD error for training $k$WTA network. These results were initially reported at \cite{Rafati-Noelle:2015:CogSci}. }
\label{plots:puddleworld}
\end{figure*}

Figure~\ref{plots:puddleworld}(a)-(c) show the learned value function (plotted as
$\max_{a}{Q}(s,a)$ for each location, $s=(x,y)$, for
representative networks of each kind. Also, Figure ~\ref{plots:puddleworld}(d)-(f) display the learned policy at the end of learning. Finally, we show learning curves displaying the episode average value of the TD Error, $\delta$, over episodes in Figure ~\ref{plots:puddleworld}(g)-(i).

In general, the Linear network did not consistently learn to solve this
problem, sometimes failing to reach the goal or choosing paths through
puddles. The Regular backpropagation network performed much better, but its
value function approximation still contained sufficient error to
produce a few poor action choices. The $k$WTA network, in comparison,
consistently converged on good approximations of the value function,
almost always resulting in optimal paths to the goal.

\begin{figure}[hbt!]
\centering
\includegraphics[width=0.7\textwidth]{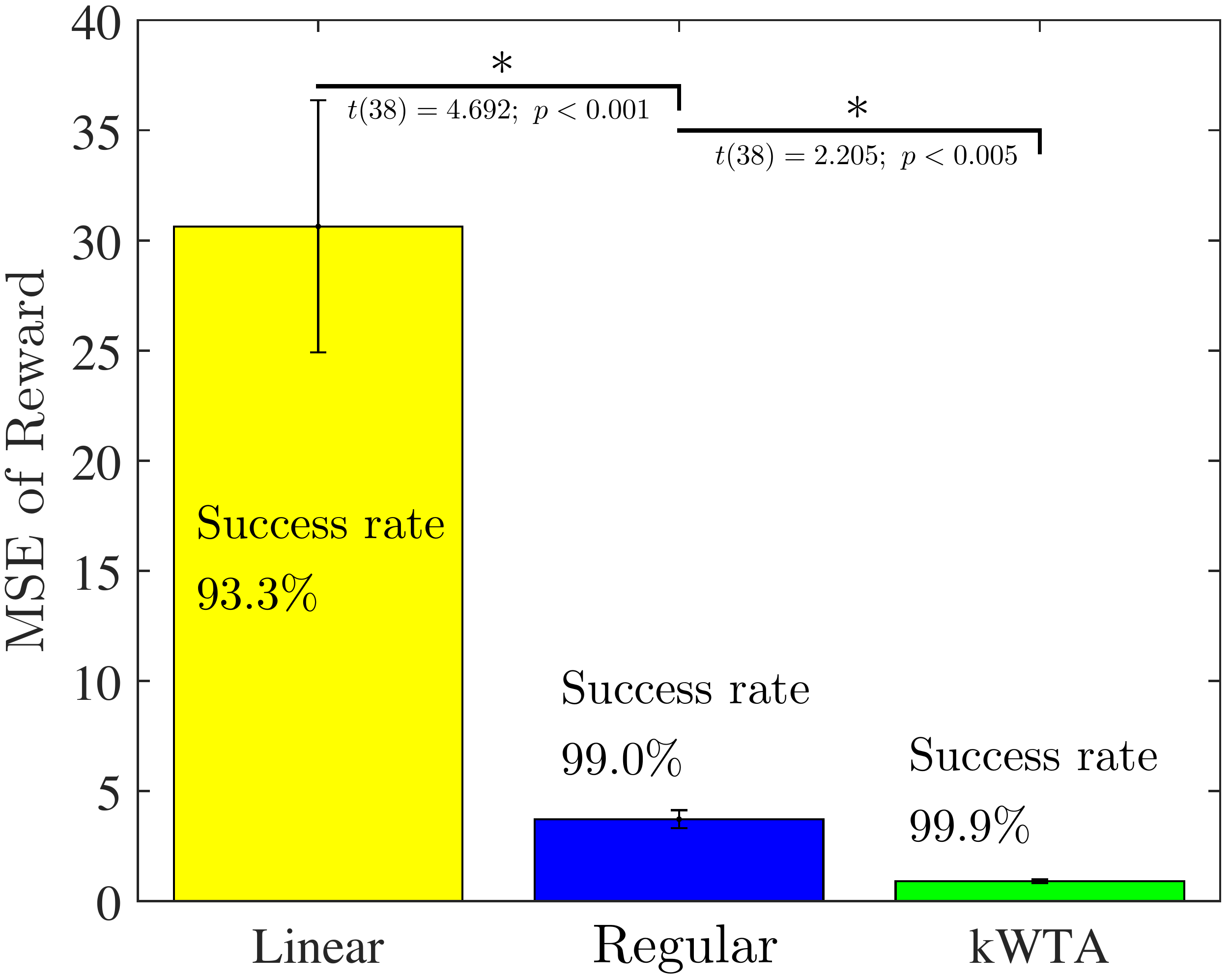}
\caption{Averaged over $20$ simulations of each network type, these
		columns display the mean squared deviation of accumulated
		reward from that of optimal performance. Error bars show one
		standard error of the mean \citep{Rafati-Noelle:2015:CogSci}.}
\label{plot:mse-plot}
\end{figure}

For each network, we quantitatively assessed the quality of the paths
produced by agents using the learned value function
approximations. For each simulation, we froze the connection weights
(or, equivalently, set the learning rate to zero), and we sequentially
produced an episode for each possible starting location, except for
locations inside of the puddles. The reward accumulated over each
episode was recorded, and, for each episode that ended at the goal
location, we calculated the sum squared deviation of these accumulated
reward values from that produced by an optimal agent (identified using
SARSA with a large look-up table for the value function). The mean of
this squared deviation measure, over all successful episodes, was
recorded for each of the $20$ simulations run for each of the $3$
network types. The means of these values, over $20$ simulations, are
displayed in Figure~\ref{plot:mse-plot}. The backpropagation network had
significantly less error than the linear network ($t(38) = 4.692$; $p
< 0.001$), and the $k$WTA network had significantly less error than the
standard backpropagation network ($t(38) = 6.663$; $p < 0.001$). On
average, the $k$WTA network deviated from optimal performance by less
than one reward point.

We also recorded the fraction of episodes which succeeded at reaching
the goal for each simulation run. The mean rate of goal attainment,
across all non-puddle starting locations, for the linear network, the 
backpropagation network, and the $k$WTA network were $93.3\%$, $99.0\%$,
and $99.9\%$, respectively. Despite these consistently high success
rates, the linear network exhibited significantly more failures than
the backpropagation network ($t(38) = 2.306$; $p < 0.05$), and the
backpropagation network exhibited significantly more failures than the
$k$WTA network ($t(38) = 2.205$; $p < 0.05$).

\subsection{The mountain-car task}
\begin{figure*}[hbt!]
\centering
\begin{tabular}{lccc}
	{}& \raisebox{5mm}{Linear} & \raisebox{5mm}{Regular}  & \raisebox{5mm}{$k$WTA} \\
	\rotatebox{90}{Values} &
	\includegraphics[width=0.28\textwidth]{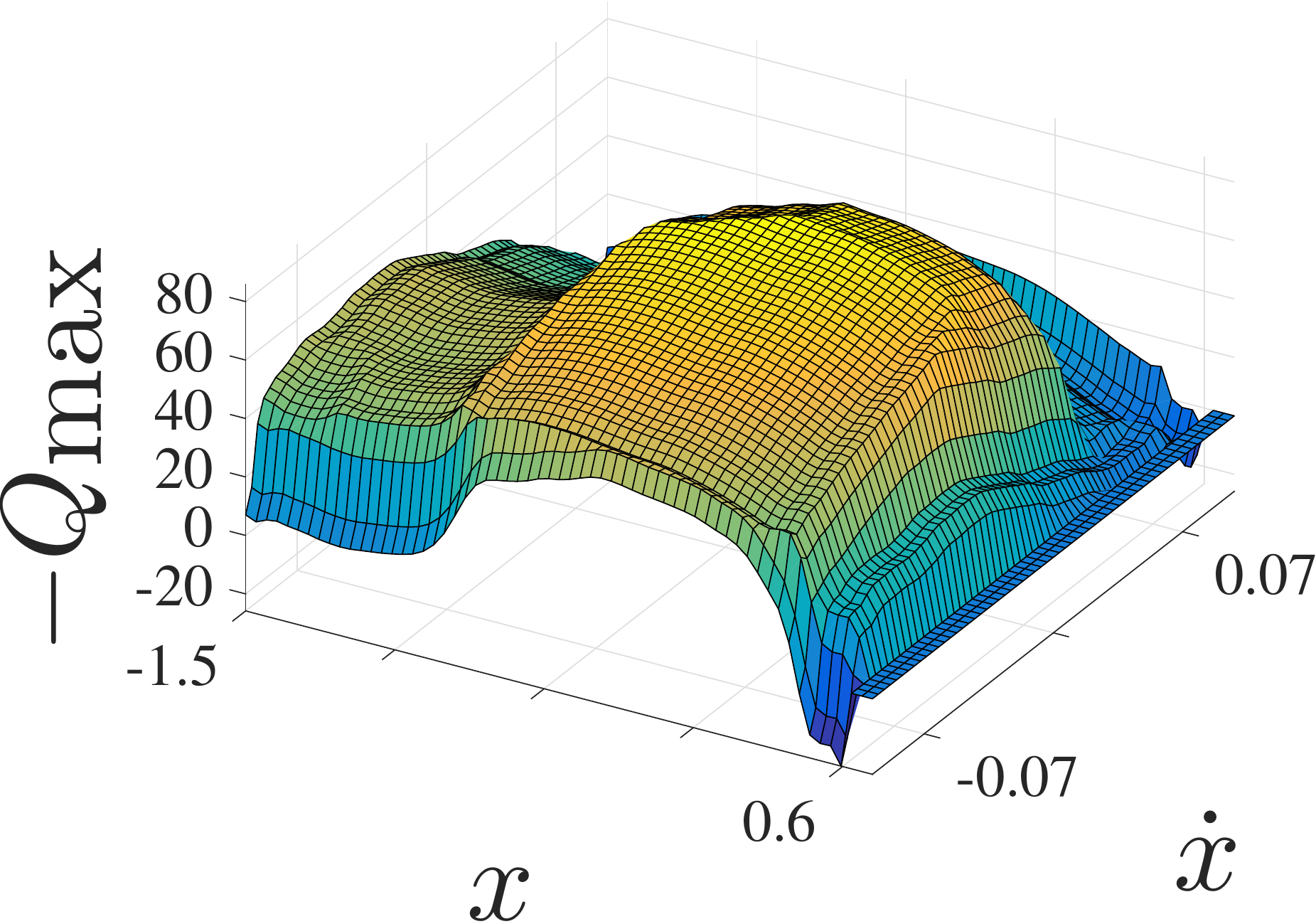} &
	\includegraphics[width=0.28\textwidth]{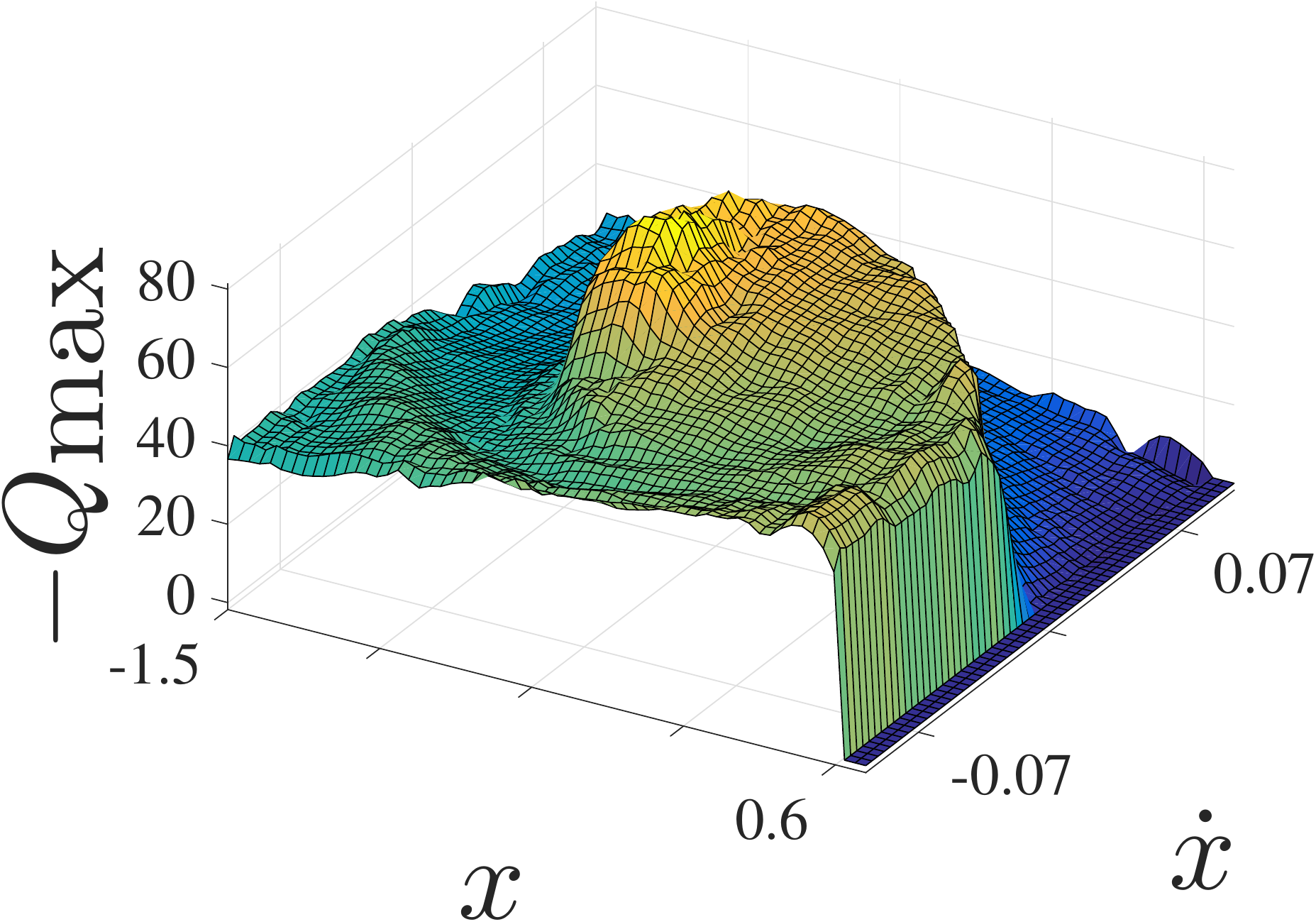} & 
	\includegraphics[width=0.28\textwidth]{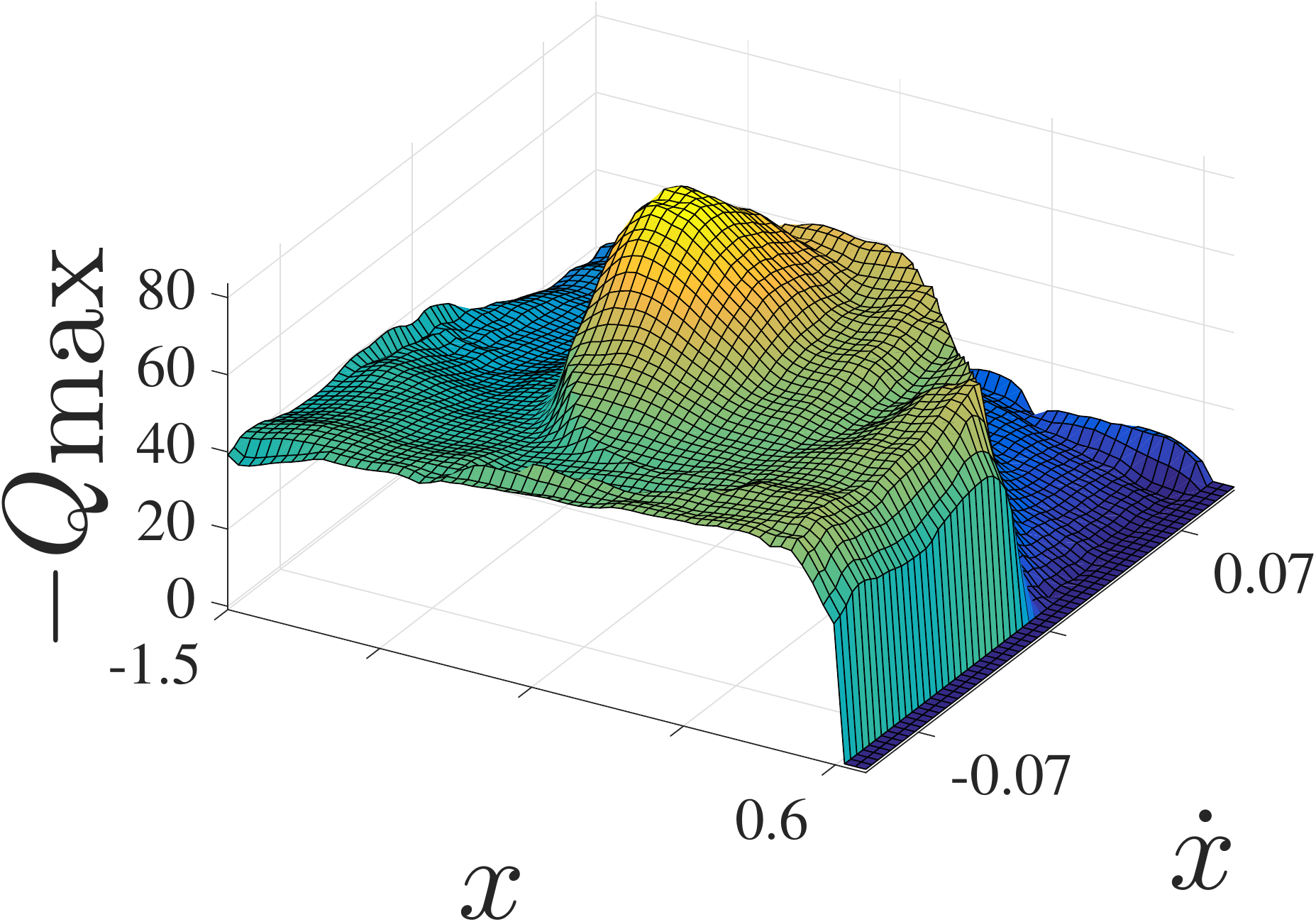}\\
	& (a) & (b) & (c) \vspace{0.1cm} \\
	\rotatebox{90}{\footnotesize\, Train performance} &
	\includegraphics[width=0.28\textwidth]{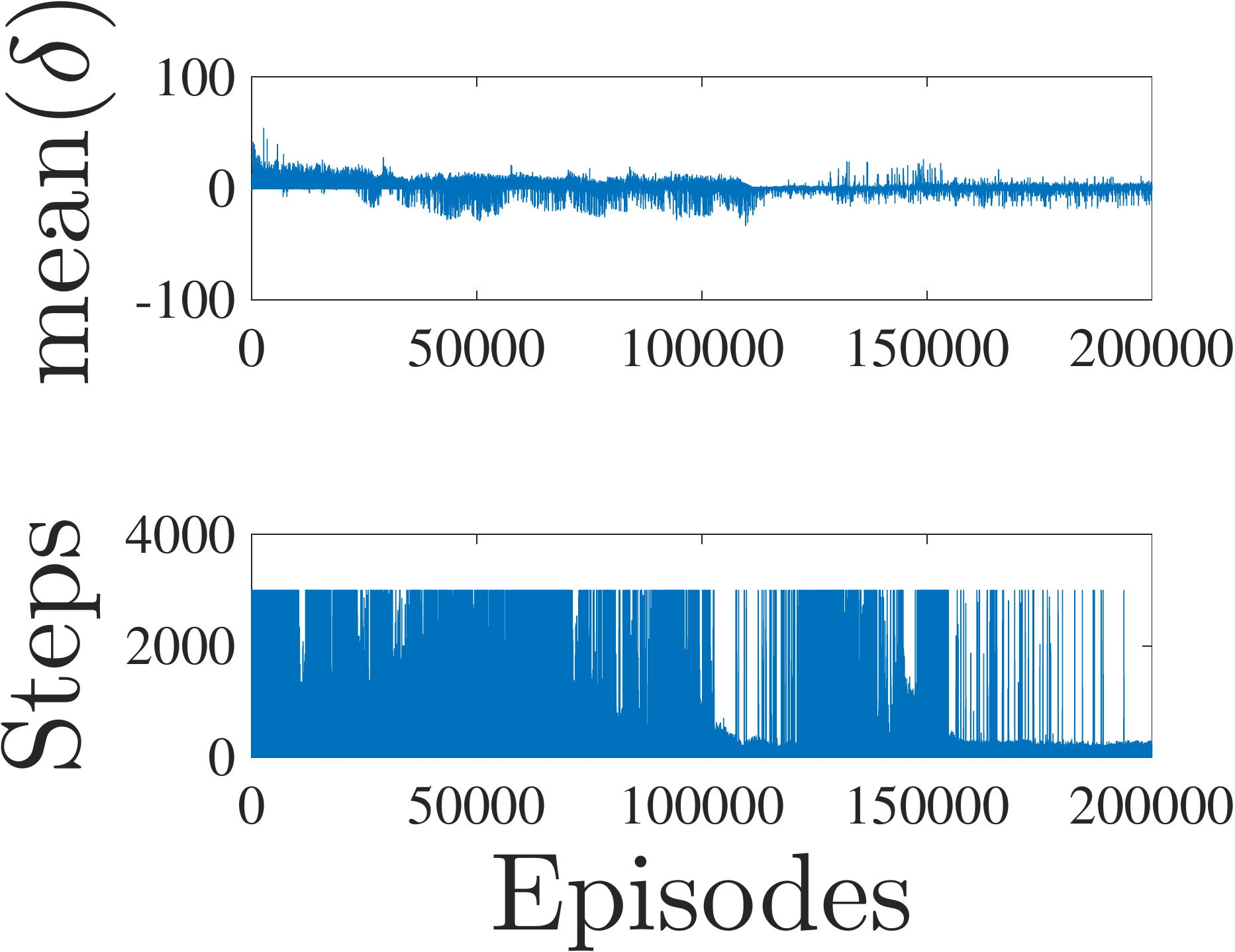} &
	\includegraphics[width=0.28\textwidth]{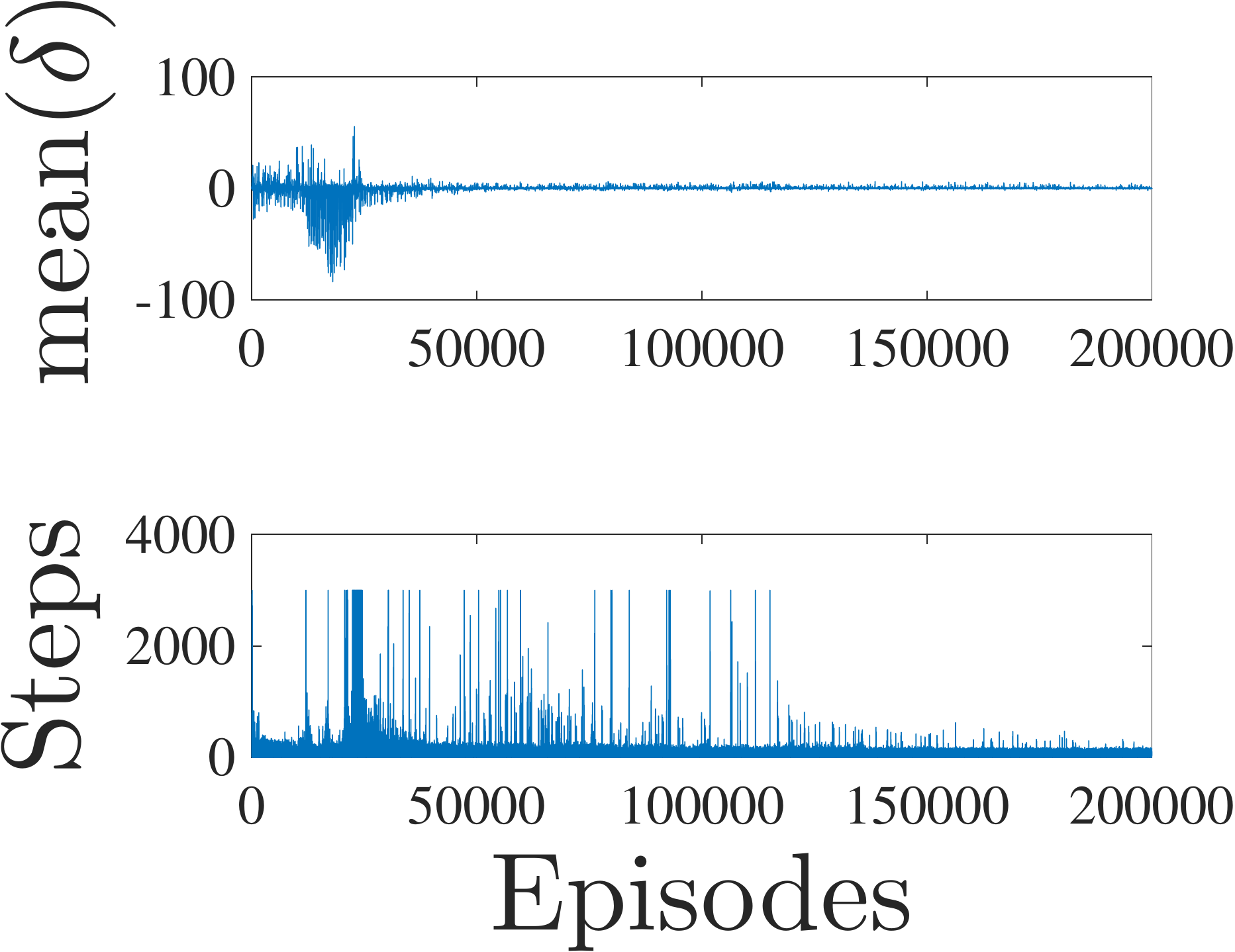} & 
	\includegraphics[width=0.28\textwidth]{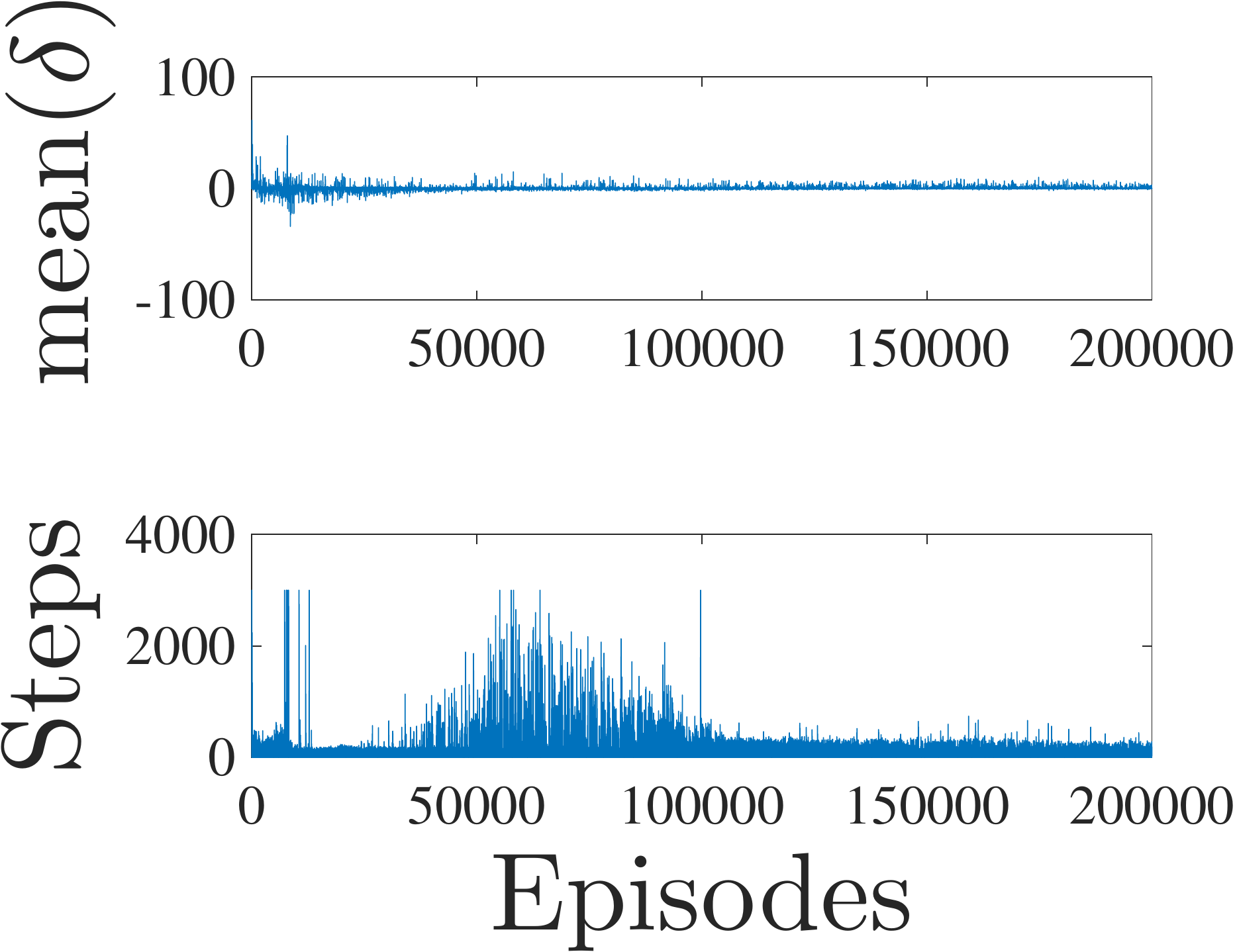}\\
	& (d) & (e) & (f) \vspace{0.1cm} \\
	\rotatebox{90}{\footnotesize\,Test performance} &
	\includegraphics[width=0.28\textwidth]{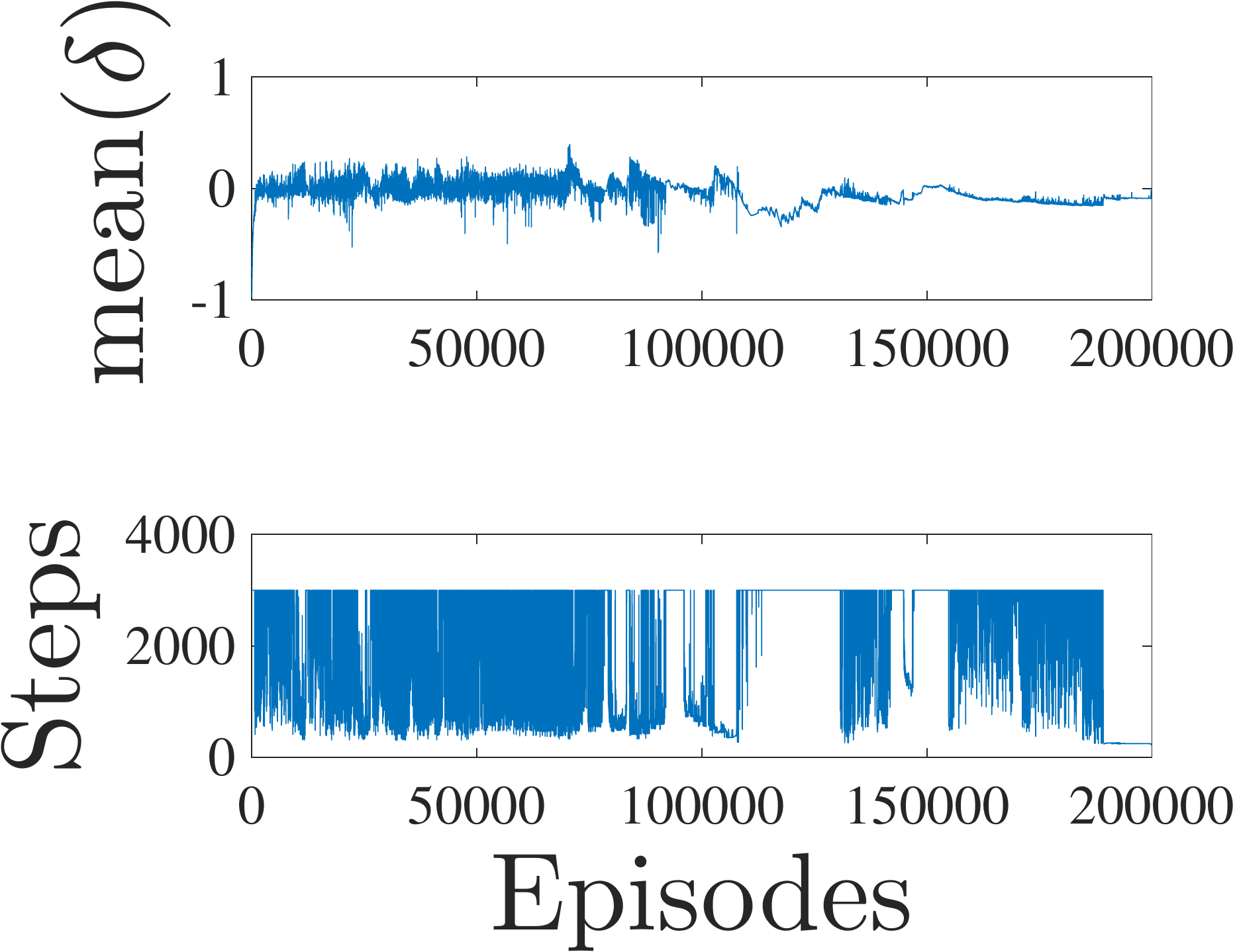}  &
	\includegraphics[width=0.28\textwidth]{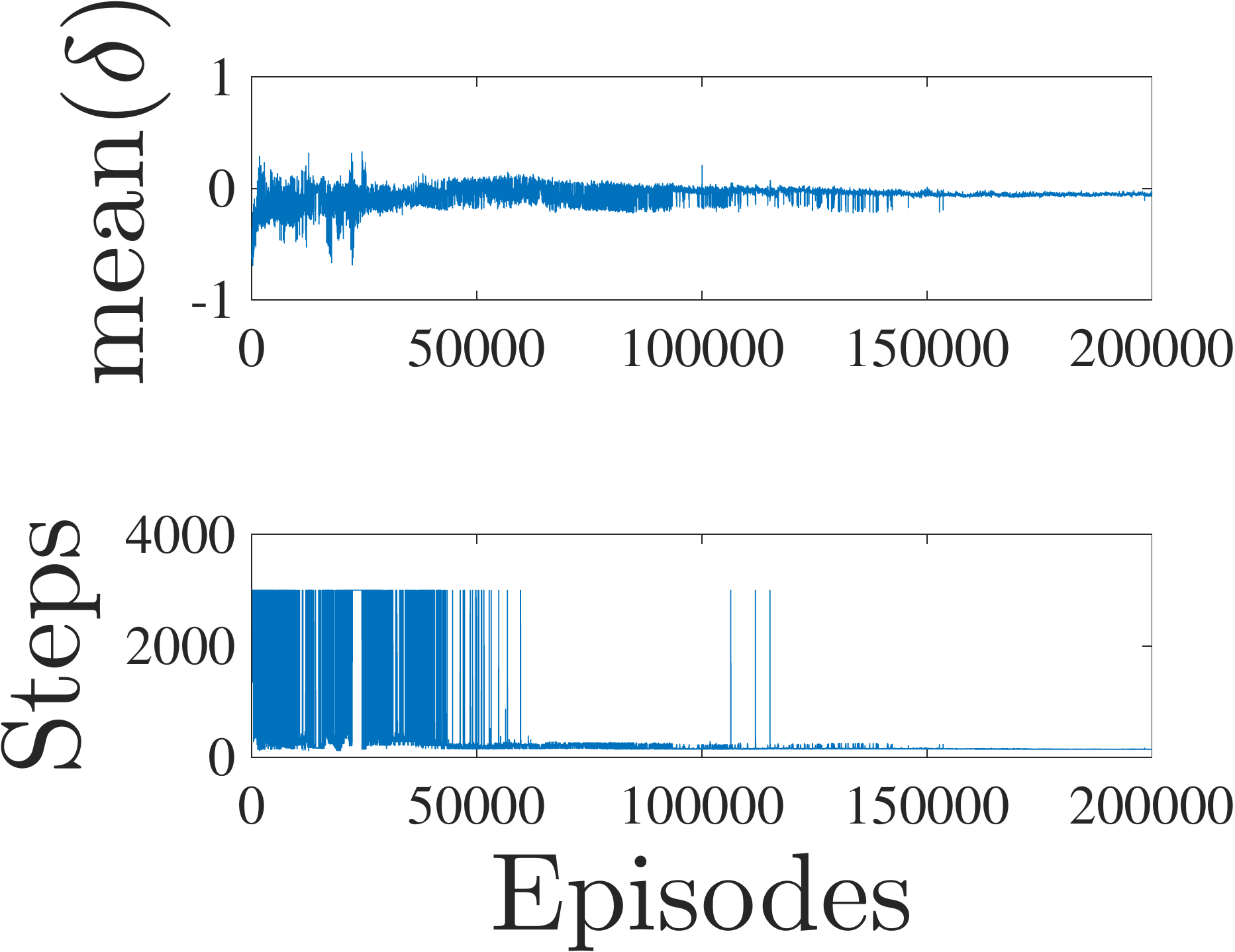} &
	\includegraphics[width=0.28\textwidth]{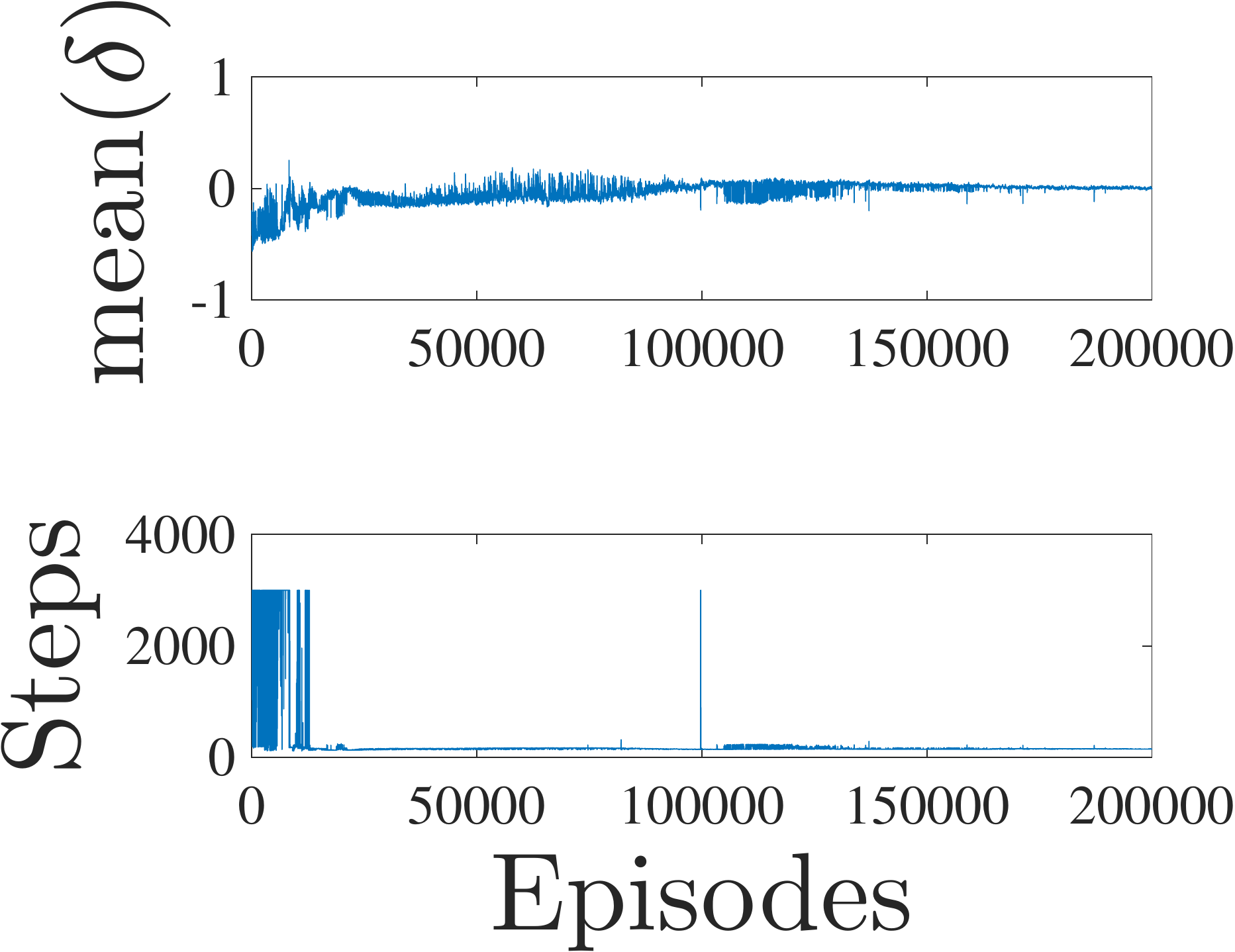}\\
	& (g) & (h) & (i)    
\end{tabular}
\caption{The performance of various networks trained to perform the
		Mountain-car task. The top row contains the approximate value of
		states after training ($\max_a {Q}(s,a)$). (a) Values approximated from Linear network. (b) Values approximated from Regular BP network. (c) Values approximated from the $k$WTA network. The middle row shows two statistics over training episodes: the top subplot shows the average value of the TD Error, $\delta$, and the bottom subplot shows the number of time steps during training episodes. Average of TD Error and total steps in training for (d) Linear (e) Regular (f) $k$WTA is given. The last row reports the testing performance, which was measured after each epoch of 1000 training episodes. During the test episodes the weight parameters were frozen and no exploration was allowed. Average of TD Error and the total steps for (g) Linear (h) Regular (i) $k$WTA are given.} 
	\label{plots:mountain-car}
\end{figure*}

Figure~\ref{plots:mountain-car} (a)-(c) show the value function (plotted as
$\max_{a} Q(s,a)$ for each location, $s$) for
representative networks of each kind. In the middle row of Figure~\ref{plots:mountain-car} (d)-(f), we show the learning curves, displaying the episode average value of the TD Error, $\delta$, over training episodes, and also the number of time steps needed to reach the goal during training episodes. In the last row, we display the results of testing the performance of the networks. The test performances were collected after every epoch of 1000 training episodes, and there were no changes to the weights and no exploration during the testing phase. The value function for the $k$WTA network is the closest numerically to optimal $Q$-table results, and the policy after training for the $k$WTA network is the most stable.

\subsection{The Acrobot task}
\begin{figure*}[hbt!]
\centering
\begin{tabular}{lccc}
	& {Linear} & {Regular}  & {$k$WTA} \\
	{\rotatebox{90}{\footnotesize Train performance}} &
	\includegraphics[width=0.28\textwidth]{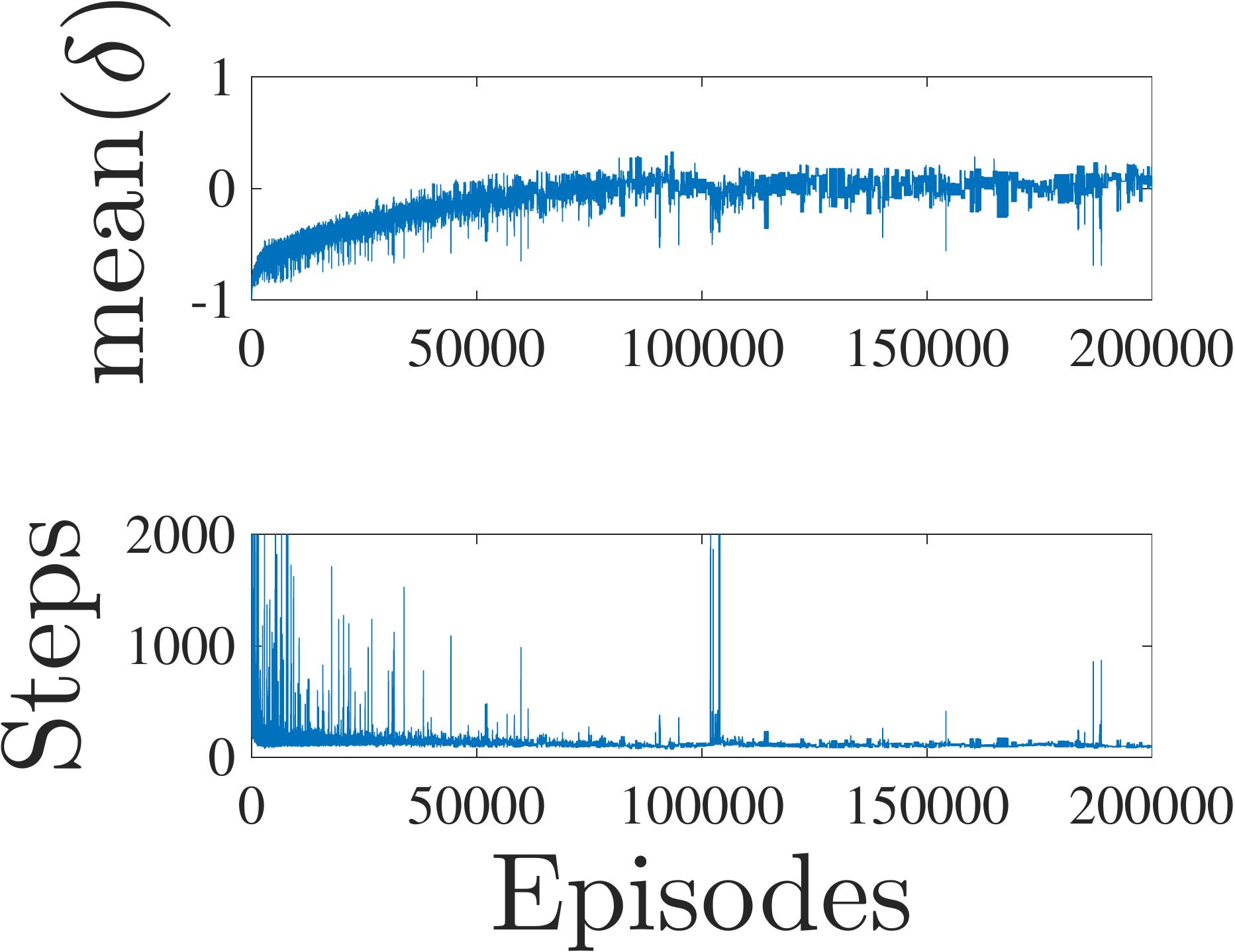}  & \includegraphics[width=0.28\textwidth]{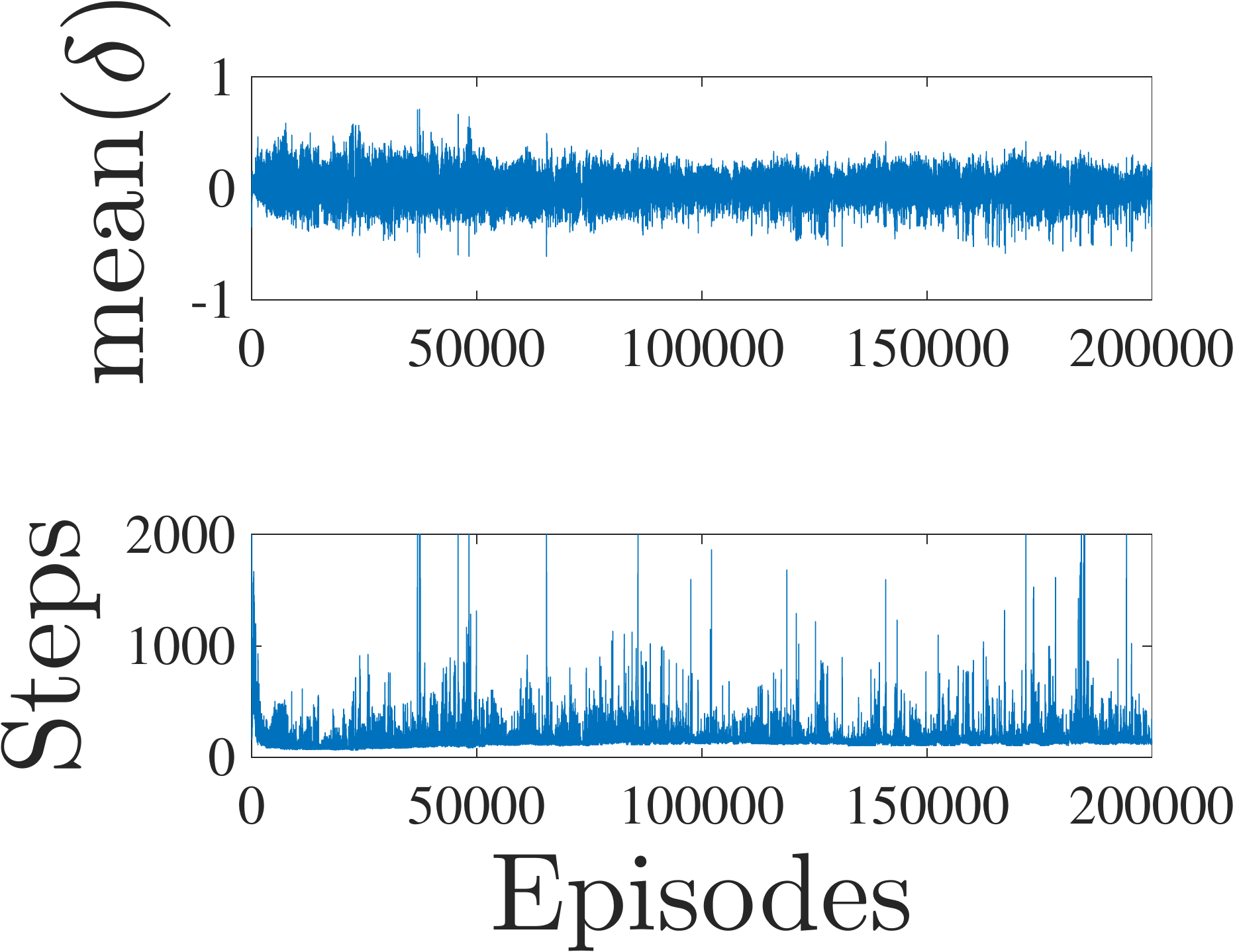} & \includegraphics[width=0.28\textwidth]{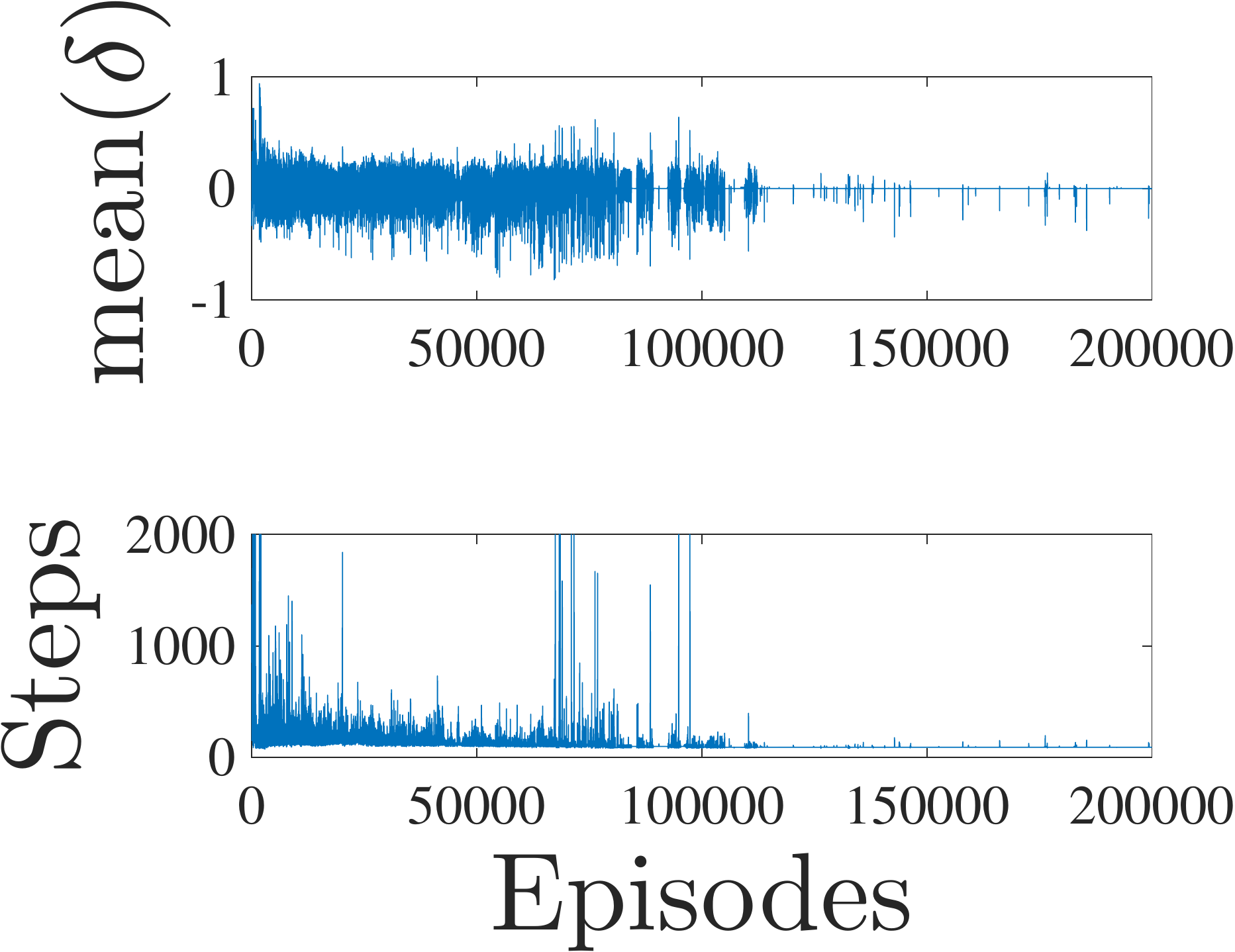} \\
	& (a) & (b) & (c) \vspace{0.1cm}\\
	{\rotatebox{90}{\footnotesize Test performance}} & \includegraphics[width=0.28\textwidth]{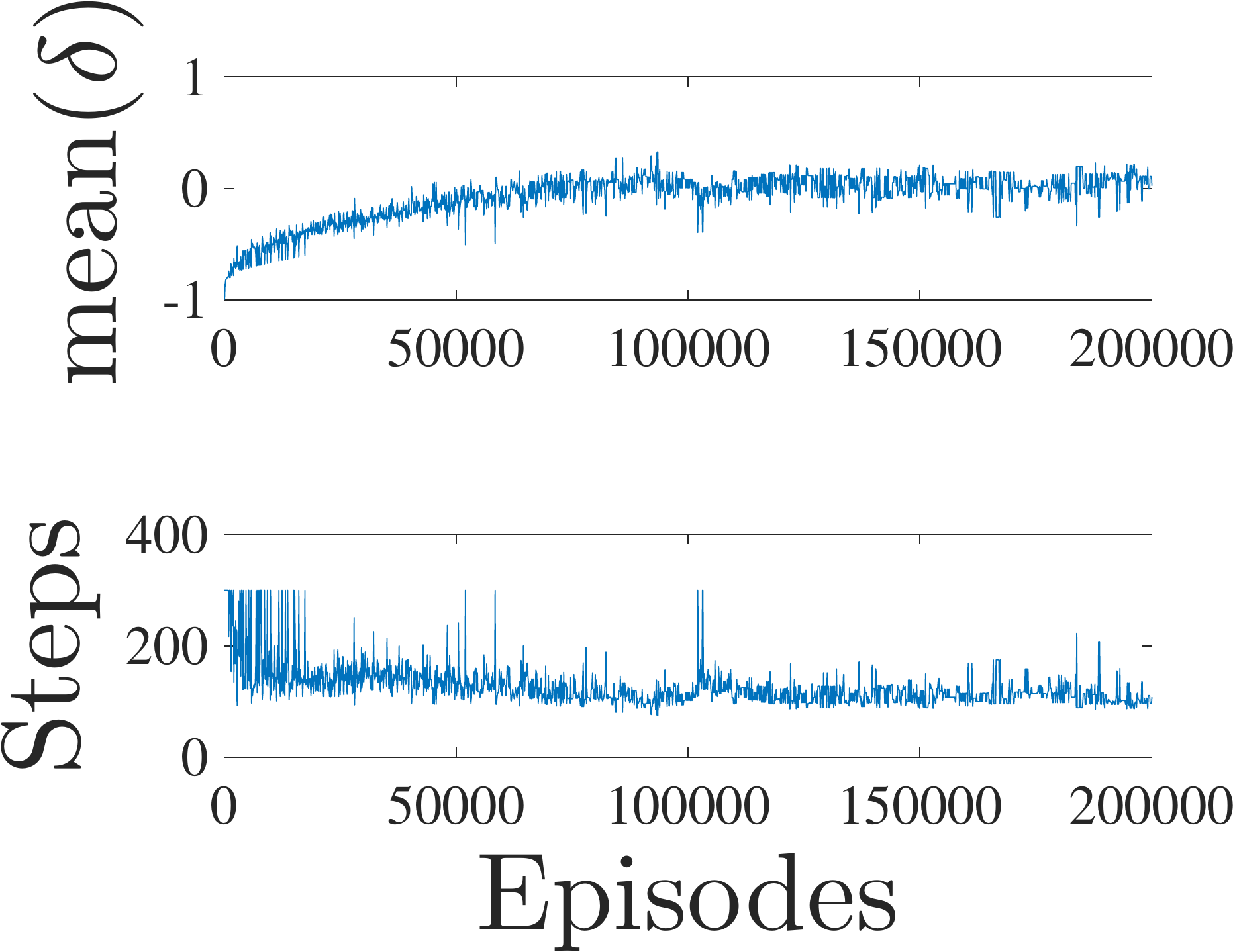} &   
	\includegraphics[width=0.28\textwidth]{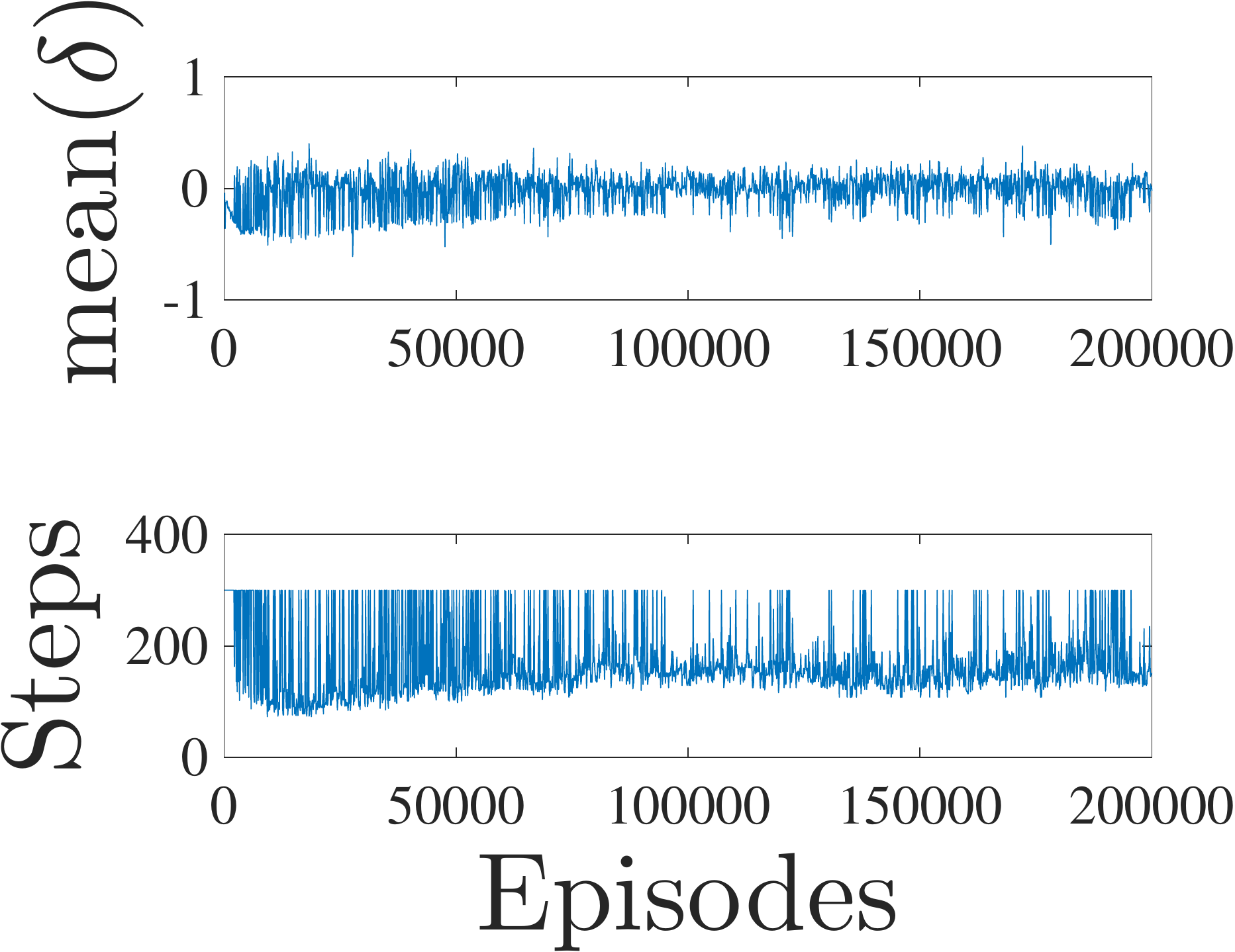} &
	\includegraphics[width=0.28\textwidth]{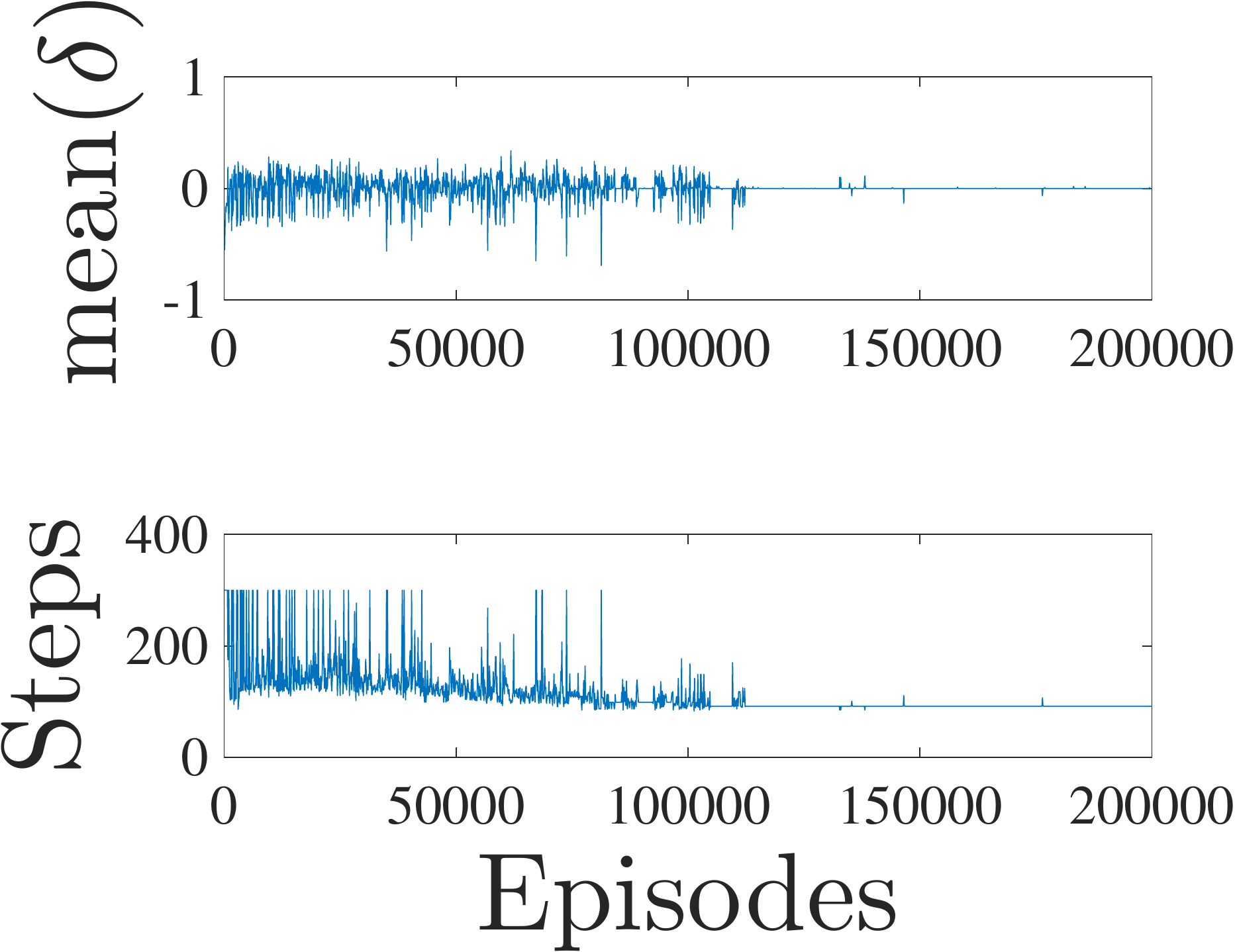} \\
	& (d) & (e) & (f) 
\end{tabular}
\caption{The performance of various networks trained to solve the
		Acrobot control task. The top row results correspond to training performance of the SARSA Algorithm \ref{Algo:SARSA}. The average TD Error (top subplot) and the total number of steps (bottom subplot) for each episode of learning are given for (a) Linear (b) Regular and (c) $k$WTA neural networks are given. The bottom row are the performance results for the testing that was measured after every epoch of 1000 training episodes in which the weight parameters were frozen and no exploration was allowed. Average of TD error and total steps for (d) Linear (e) Regular (f) $k$WTA are given.}
\label{plots:acrobot}
\end{figure*}

In the first row of Figure~\ref{plots:acrobot}(a)-(c), the episode average value of the TD Error, $\delta$, over training episodes, and also the number of time steps
needed to reach the goal during training episodes are shown for different networks, during the training phase. In the last row of Figure~\ref{plots:acrobot}(d)-(f), we display the results of testing performance of the networks for the three network architectures. At testing time, any changes to the weights were avoided, and there was no exploration. 

From these results, we can see that only the $k$WTA network could learn the optimal policy. Both the Linear network and the Regular backpropagation network failed to learn the optimal policy for the Acrobot control task.

%% file: sections/conclusions.tex
Using a function approximator to learn the value function (an estimate of the expected future reward from a given environmental state) has the benefit of supporting generalization across similar states, but this approach can produce a form of catastrophic interference that hinders learning, mostly arising in cases in which similar states have widely different values. Computational neuroscience models have shown that a combination of feedforward and feedback inhibition in neural circuits naturally produces sparse conjunctive codes over a collection of excitatory neurons \citep{Noelle-2008:BICA}. 

Artificial neural networks can be forced to produce such sparse codes over their hidden units by including a process akin to the sort of pooled lateral inhibition that is ubiquitous in the cerebral cortex \citep{OReillyRC:2001:CECN}. We implemented a state-action value function approximator that utilizes the $k$-Winners-Take-All mechanism \citep{OReillyRC:2001:Generalization} which efficiently incorporate a kind of lateral inhibition into artificial neural network layers, driving these machine learning systems to produce sparse conjunctive internal representations \citep{Rafati-Noelle:2015:CogSci,Rafati-Noelle:2017:CCN}.

The proposed method solves the previously impossible-to-solve control tasks by balancing between generalization (pattern completion) and sparsity (pattern separation). We produced computational simulation results as preliminary evidence that learning such sparse representations of the state of an RL agent can help compensate for weaknesses of artificial neural networks in TD Learning. 

These simulation results both support our method for improving
representation learning in model-free RL and also lend preliminary support to the hypothesis that the midbrain dopamine system, along with associated circuits in the basal ganglia, do, indeed, implement a form of TD Learning, and the observed problems with TD learning do not arise in the brain due to the encoding of sensory state information in circuits that make use of lateral inhibition.

Using a sparse conjunctive representation of the agent's state not only can help in the solving of simple reinforcement learning task, but it might also help improve the learning of some large-scale tasks too. For example, FeUdal (state-goal) networks \citep{vezhnevets2017feudal} can benefit using $k$WTA mechanism for learning sparse representations in conjunction with Hierarchical Reinforcement Learning (HRL) (see  \cite{Rafati2019phd,Rafati-Noelle:2019:HRL-arXiv,Rafati-Noelle:2019:AAAI-KEG,Rafati-Noelle:2019:SPiRL}). In the future, we will extend this work to the deep RL framework, where the value function is approximated by a deep Convolutional Neural Network (CNN). The $k$WTA mechanism can be used in the fully connected layers of the CNN in order to generate sparse representations using the lateral inhibition like mechanism.